\documentclass{article} % For LaTeX2e
\usepackage[final]{colm2026_conference}

\usepackage{microtype}
\usepackage{hyperref}
\usepackage{url}
\usepackage{booktabs}
\usepackage{algorithm}
\usepackage{amsmath}
\usepackage{wrapfig}
\usepackage{graphicx}
\usepackage{multicol}
\usepackage{multirow}
\usepackage{subcaption}
\usepackage{algorithmic}
\usepackage{amssymb}
\usepackage[table]{xcolor}
\usepackage{pifont}
\DeclareMathAlphabet\mathbfcal{OMS}{cmsy}{b}{n}

\newcommand{\mat}[1]{\mathbf{#1}}

\usepackage{lineno}

\definecolor{darkblue}{rgb}{0, 0, 0.5}
\hypersetup{colorlinks=true, citecolor=darkblue, linkcolor=darkblue, urlcolor=darkblue}

\title{MuonQ: Enhancing Low-Bit Muon Quantization via Directional Fidelity Optimization}

\author{Yupeng Su, Ruijie Zhang, Ziyue Liu, Yequan Zhao, Zheng Zhang \\
University of California, Santa Barbara \\
\texttt{\{yupengsu, zzhang01\}@ucsb.edu}\\
}

\begin{document}

\ifcolmsubmission
\linenumbers
\fi

\maketitle

\begin{abstract}
The Muon optimizer has emerged as a compelling alternative to Adam for training large language models, achieving remarkable computational savings through gradient orthogonalization. However, Muon's optimizer state is more sensitive to quantization errors: because the orthogonalization discards the magnitudes of singular values and retains only directional information, even small quantization errors in singular vector directions are amplified in the update. In this work, we propose MuonQ, a low-bit Muon training framework built on the principle of directional fidelity optimization. First, we apply a pre-quantization normalization so that each step introduces quantization errors of the same magnitude, preventing the accumulated error from developing a preferred direction. Second, we introduce a structural decomposition that separately quantizes the dominant singular components via power iteration, ensuring that quantization errors perturb only singular value magnitudes rather than rotating singular vector directions. Third, we adopt $\mu$-law companding quantization to allocate higher resolution to densely packed momentum values, shifting the quantization objective from outlier preservation to dense-region distinguishability. Together, these techniques enable stable 4-bit quantization of Muon's optimizer states. {Pre-training experiments on GPT-style and LLaMA-style models demonstrate that MuonQ at 4-bit precision recovers most of full-precision Muon's training loss and downstream accuracy while reducing optimizer-state memory by up to 7.3$\times$, offering a favorable point on the accuracy--memory trade-off.} Our code is available at \url{https://github.com/YupengSu/MuonQ}. 
\end{abstract}

\section{Introduction}
\label{sec:intro}
Large language models (LLMs) have made transformative impacts across a broad spectrum of domains, including natural language understanding, code generation, and mathematical reasoning. However, as model scale increases following established scaling laws~\citep{kaplan2020scaling,hoffmann2022training}, the memory required for training has become a critical bottleneck. Modern adaptive optimizers such as Adam~\citep{kingma2015adam} and AdamW~\citep{loshchilov2019decoupled} maintain first- and second-order moment estimates for every trainable parameter, resulting in a memory overhead of at least twice the size of the model itself. A growing body of work has demonstrated that these states can be aggressively compressed~\citep{dettmers2022optimizers,wang2024fourbit,zhang2024qgalore,xu2025solo}, without significant degradation in training quality.

Meanwhile, the Muon optimizer~\citep{jordan2024muon} has emerged as a compelling alternative to Adam for LLM training.
By orthogonalizing the gradient momentum via Newton-Schulz iterations, Muon achieves spectrally controlled updates that yield faster convergence and up to $2\times$ computational savings over AdamW, as validated at scales reaching 16 billion parameters~\citep{liu2025muon}.
A growing body of work has also begun to investigate the quantization of Muon's optimizer states. \citet{gupta2025quantmuon} showed that 8-bit blockwise quantization preserves Muon's performance reliably, and theoretical analysis has further suggested that Muon is inherently more robust to quantization than Adam~\citep{tang2026convergence}. However, pushing to 4-bit remains challenging: naive uniform quantization leads to significant degradation, and current approaches resort to mixed-precision schemes that keep the dominant singular components at higher precision~\citep{wu2026lowbit}.

% \zz{What is the goal of this paper? Save memory or speed up training? Memory saving is less important than training speedup.} 
In this work, we propose MuonQ, a low-bit Muon training framework that enables pure 4-bit quantization via directional fidelity optimization. Prior quantization methods for Muon focus primarily on numerical fidelity, minimizing the reconstruction error of the momentum. However, we observe that Muon's polar decomposition discards singular value magnitudes and retains only directional information, so quantization errors that perturb singular vector directions are fully preserved in the update. The appropriate objective is therefore to preserve the directional structure of the momentum rather than its element-wise accuracy. Guided by this principle, we make the following contributions:

\begin{enumerate}
    \item \textbf{Pre-quantization normalization for uniform error accumulation} (\S\ref{sec:method-normalize}). We normalize gradients and momentum before quantization so that each step introduces quantization errors of the same magnitude, preventing the accumulated error from developing a preferred direction.
    
    \item \textbf{Structural decomposition for stable orthogonalization} (\S\ref{sec:method-subspace}). We show that orthogonalization amplifies quantization errors. By decomposing the momentum and quantizing the singular factors independently, we ensure that quantization errors perturb only singular value magnitudes rather than rotating singular vector directions. A truncated top-$k$ variant further balances memory and accuracy.

    \item \textbf{Companding quantization for dense-region distinguishability} (\S\ref{sec:method-companding}). Muon requires high resolution near zero due to equal weighting of singular directions. We apply $\mu$-law companding to reallocate quantization bins toward this dense region.
\end{enumerate}
Together, these techniques enable stable pure 4-bit quantization of Muon's optimizer states. 

\section{Background}
% \zz{Please check the paper writing guideline on my research homepage regarding font types for different kinds of variables (scalars, vectors/matrices and tensors). } (syp: fix it)

\paragraph{Muon Optimizer.}
\label{sec:background_muon}
The Muon optimizer~\citep{jordan2024muon} is designed for matrix-shaped parameters in neural network hidden layers.
Unlike Adam, which maintains per-element first and second moment estimates, Muon applies a matrix-level orthogonalization to the gradient momentum, producing spectrally controlled updates.
Specifically, for a matrix-shaped parameter $\mat{W} \in \mathbb{R}^{m \times n}$, let $\mat{G}_t = \nabla_\mat{W} \mathcal{L}_t$ denote the gradient at step $t$. Muon maintains a momentum buffer $\mat{M}_t$ and computes updates through the following procedure:
\begin{equation}
    \mat{M}_t = \beta \, \mat{M}_{t-1} + \mat{G}_t, \quad \mat{W}_{t} = \mat{W}_{t-1} - \eta \, \mathrm{polar}(\mat{M}_t),
\end{equation}
where $\beta$ is the momentum coefficient, $\eta$ is the learning rate, and $\mathrm{polar}(\cdot)$ denotes the orthogonal polar factor of a matrix. For a matrix with singular value decomposition $\mat{M}_t = \mat{U} \mat{\Sigma} \mat{V}^\top$, this is defined as $\mathrm{polar}(\mat{M}_t) = \mat{U}\mat{V}^\top$. The polar factor replaces all singular values of the momentum with ones, yielding a spectrally flat update that prevents any single direction from dominating the optimization trajectory~\citep{bernstein2024old}.
In practice, $\mathrm{polar}(\cdot)$ is efficiently approximated via Newton--Schulz iterations or improved variants such as Polar Express~\citep{amsel2026polar}. 
% To scale Muon to LLM training, \citet{liu2025muon} further introduced weight decay and per-parameter update scaling, we adopt their protocol throughout our experiments.
% Following prior work~\citep{bernstein2025deriving,liu2025muon}, we apply a dimensional pre-factor $\sqrt{m/n}$ to the update and rescale the learning rate by $0.2 *\max(m, n)$, which improves scalability across model sizes and allows Muon and AdamW to share a common learning rate schedule for the co-optimized parameter groups.

\paragraph{Quantization and Dequantization.}
\label{sec:background_quantization}
Given an input tensor $\mat{x}$, the standard $b$-bit symmetric uniform quantization at granularity $g$ partitions $\mat{x}$ into groups and computes a per-group scale factor and integer codes as:
\begin{equation}
    \mat{s} = \frac{\max(|\mat{x}|)}{2^{b-1}-1}, \quad
    \mat{q} = \mathrm{clamp}\!\left(
        \mathrm{round}\!\left(\frac{\mat{x}}{\mat{s}}\right),
        -2^{b-1}+1,\;
        2^{b-1}-1
    \right),
\end{equation}
where $\mat{s}$ is the scale factor computed over each granularity group and $\mat{q} \in \mathbb{Z}^{|\mat{x}|}$ is the corresponding integer code. The dequantization operator reconstructs the approximation:
\begin{equation}
    \hat{\mat{x}} = \mathrm{Dequant}(\mat{q}, \mat{s}) = \mat{q} \cdot \mat{s}.
\end{equation}
We denote the full quantization pipeline as $\mat{q}, \mat{s} = \mathrm{Quant}_b^g(\mat{x})$.

\begin{algorithm}[tbp]
\caption{A Single Training Step of MuonQ}
\label{alg:qmuon}
\begin{algorithmic}[1]
\REQUIRE Gradient $\mat{G}_t$, learning rate $\eta$, momentum $\beta$, bit-width $b$, rank $k$, parameters $\mat{W}_{t-1}$
\REQUIRE Companding quantizer $\mathrm{CQuant}_b^{g}$ and dequantizer $\mathrm{CDequant}$ (Eq.~\ref{eq:companding-quant})
\REQUIRE Optimizer states $(\mat{U}_{t-1}^q, \mat{U}_{t-1}^s),\; (\mat{S}_{t-1}^q, \mat{S}_{t-1}^s),\; (\mat{R}_{t-1}^q, \mat{R}_{t-1}^s)$; or $\varnothing$ if $t=1$
\IF{$t = 1$}
    \STATE $\hat{\mat{M}}_{t-1} \leftarrow \mathbf{0} \in \mathbb{R}^{m \times n}$;\quad $\hat{\mat{S}}_{t-1} \leftarrow \mathcal{N}(\mat{0}, \mat{I}) \in \mathbb{R}^{k \times n}$ \hfill $\triangleright$ Random initialization
\ELSE
    \STATE $\hat{\mat{U}}_{t-1},\; \hat{\mat{S}}_{t-1} \leftarrow \mathrm{CDequant}(\mat{U}_{t-1}^q, \mat{U}_{t-1}^s),\; \mathrm{CDequant}(\mat{S}_{t-1}^q, \mat{S}_{t-1}^s)$
    \STATE $\hat{\mat{R}}_{t-1} \leftarrow \mathrm{CDequant}(\mat{R}_{t-1}^q, \mat{R}_{t-1}^s)$
    \STATE $\hat{\mat{M}}_{t-1} \leftarrow \hat{\mat{U}}_{t-1} \cdot \hat{\mat{S}}_{t-1} + \hat{\mat{R}}_{t-1}$ \hfill $\triangleright$ Reconstruct momentum
\ENDIF
\STATE {\textit{//} Pre-quantization normalization (\S\ref{sec:method-normalize})}
\STATE $\mat{M}_t \leftarrow \beta \, \hat{\mat{M}}_{t-1} + \mat{G}_t \,/\, \|\mat{G}_t\|_F$
\STATE $\bar{\mat{M}}_t \leftarrow \mat{M}_t \,/\, \|\mat{M}_t\|_F$
\STATE {\textit{//} Structural decomposition (\S\ref{sec:method-subspace})}
\STATE $\mat{V}_{t-1} \leftarrow \mathrm{RowNorm}(\hat{\mat{S}}_{t-1})$
\hfill $\triangleright$ Row-wise normalization
\STATE $ \mat{U}_t \leftarrow \mathrm{orth}(\bar{\mat{M}}_t \, \mat{V}_{t-1}^{\!\top}),\quad \mat{S}_t \leftarrow \mat{U}_t^{\!\top}\,\bar{\mat{M}}_t, \quad \mat{R}_t \leftarrow \bar{\mat{M}}_t - \mat{U}_t \, \mat{S}_t$
\hfill $\triangleright$ Power iteration
\STATE {\textit{//} Companding quantization (\S\ref{sec:method-companding})}
\STATE $(\mat{U}_t^q, \mat{U}_t^s),\; (\mat{S}_t^q, \mat{S}_t^s) \leftarrow \mathrm{CQuant}_b^{\mathrm{col}}(\mat{U}_t),\; \mathrm{CQuant}_b^{\mathrm{row}}(\mat{S}_t)$
\STATE $(\mat{R}_t^q, \mat{R}_t^s) \leftarrow \mathrm{CQuant}_b^{g}(\mat{R}_t)$
\STATE $\mat{W}_t \leftarrow \mat{W}_{t-1} - \eta \, \mathrm{polar}(\bar{\mat{M}}_t)$ \hfill $\triangleright$ {Parameter update}
\end{algorithmic}
\end{algorithm}

\section{Methodology}
\label{sec:methodology}

In this section, we present MuonQ, a unified framework for stable pure 4-bit quantization of Muon's optimizer states, grounded in the principle of directional fidelity optimization. As discussed in Section~\ref{sec:background_muon}, Muon's polar decomposition discards singular value magnitudes and retains only directional information, making the optimizer state uniquely sensitive to directional perturbations introduced by quantization. MuonQ addresses this sensitivity through three complementary techniques: pre-quantization normalization to prevent non-uniform error accumulation (\S\ref{sec:method-normalize}), structural decomposition to ensure momentum orthogonalization stability  (\S\ref{sec:method-subspace}), and $\mu$-law companding quantization to improve resolution in the dense near-zero region (\S\ref{sec:method-companding}). Algorithm~\ref{alg:qmuon} summarizes the complete procedure, and the subsequent subsections detail each component in order.

For evaluating quantization quality, beyond the standard \textbf{relative error (RE)} that measures numerical fidelity, we additionally adopt the \textbf{cosine similarity (CS)} to directly assess directional fidelity, which is what Muon's polar decomposition ultimately relies on. Formally, for two matrices $\mat{A}$, $\mat{B}$ and Frobenius inner product $\langle \cdot, \cdot \rangle_F$, we define:
\begin{equation}
    \mathrm{RE}(\mat{A}, \mat{B}) = \frac{\|\mat{A} - \mat{B}\|_F}{\|\mat{A}\|_F}, \quad \mathrm{CS}(\mat{A}, \mat{B}) = \frac{\langle \mat{A},\, \mat{B} \rangle_F}{\|\mat{A}\|_F \cdot \|\mat{B}\|_F}.
\end{equation}

% >>> CR-NEW (camera-ready, Reviewer gTtK W3): perturbation lemma promoting CS as the derived metric
{The following bound makes precise why CS, rather than RE, is the right quantization objective for Muon: only the component of the quantization error orthogonal to $\mat{M}$ survives the polar projection. For $\mat{M}\in\mathbb{R}^{m\times n}$ with quantized reconstruction $\hat{\mat{M}}$,}
\begin{equation}
    \|\mathrm{polar}(\hat{\mat{M}}) - \mathrm{polar}(\mat{M})\|_F \;\le\; \frac{\sqrt{2}\,\|\hat{\mat{M}}\|_F}{\sigma_{\min}(\mat{M})}\sqrt{1 - \mathrm{CS}(\mat{M},\hat{\mat{M}})^2} \;+\; O\!\left(\frac{\|\mat{M}-\hat{\mat{M}}\|_F^2}{\sigma_{\min}(\mat{M})^2}\right).
\end{equation}
{Hence the update error is governed by $\sqrt{1-\mathrm{CS}^2}$, so maximizing cosine similarity---not minimizing reconstruction error---is the objective each MuonQ component targets. We present this as motivation rather than a guarantee: the constant scales as $1/\sigma_{\min}(\mat{M})$ and can be loose for ill-conditioned $\mat{M}$; the full derivation is deferred to Appendix~\ref{app:lemma-proof}.}

% ============================================================
\subsection{Pre-Quantization Normalization for Temporal Error Uniformity}
\label{sec:method-normalize}

The first step of MuonQ addresses the long-term stability of quantized training by examining how quantization errors accumulate across the momentum update $\mat{M}_t = \beta \mat{M}_{t-1} + \mat{G}_t$.

\begin{wrapfigure}{r}{0.45\textwidth}
    \centering
    \vspace{-10pt}
    \includegraphics[width=0.45\textwidth]{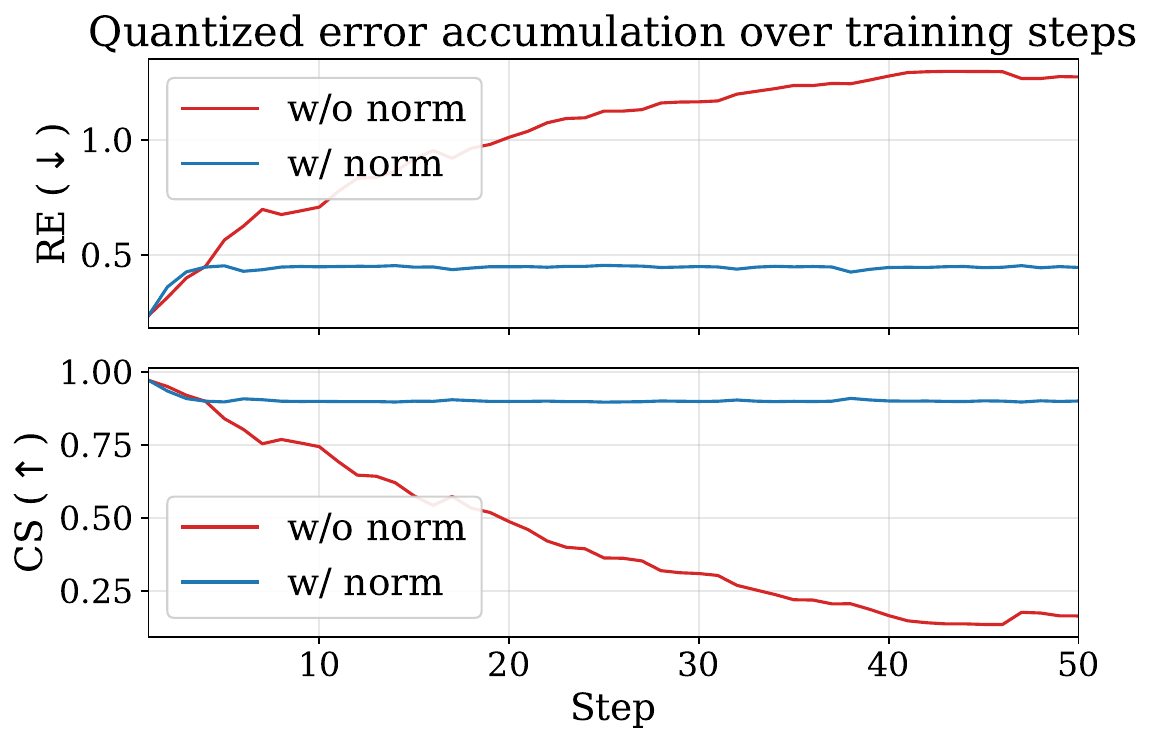}
    \caption{Quantization error accumulation over 50 momentum update steps. {Curves show the \emph{accumulated} RE and CS of the dequantized momentum against the full-precision momentum trajectory.}}
    % \zz{increase text size in the figure.}
    \label{fig:momentum-drift}
\end{wrapfigure}

\paragraph{Non-uniform error accumulation.}
In practice, the norms of $\mat{G}_t$ and $\mat{M}_{t}$ vary substantially across steps, causing the quantization error magnitude to fluctuate from step to step. When accumulated through the momentum recursion, these non-uniformly scaled errors develop a preferred direction, producing anisotropic drift. Since Muon's polar decomposition is sensitive to directional perturbations, such drift directly degrades update quality over time. As shown by the red curves in Figure~\ref{fig:momentum-drift}, without normalization, both the relative error and cosine similarity decline consistently as training progresses.

\paragraph{Pre-quantization normalization.}
We observe that if every quantization step introduces errors of the same magnitude, then the accumulated error behaves as a sum of identically scaled random perturbations, which remains approximately isotropic and does not develop a preferred direction. To achieve this, we normalize both the gradient and momentum to unit Frobenius norm \textbf{before} quantization at each step:
\begin{equation}
    \mat{M}_t = \beta \hat{\mat{M}}_{t-1} + \frac{\mat{G}_t}{\|\mat{G}_t\|_F}, \quad \bar{\mat{M}}_{t} = \frac{\mat{M}_{t}}{\|\mat{M}_{t}\|_F}.
\end{equation}
The normalized $\bar{\mat{M}}_t$ is then decomposed and quantized; upon dequantization at the next step, the reconstructed $\hat{\mat{M}}_{t-1}$ is used directly without rescaling. We verify experimentally in ablation study section (\S~\ref{app:norm-without-quant}) that applying this normalization to full-precision Muon produces training curves indistinguishable from the unnormalized baseline, confirming that the modified recursion does not alter optimization dynamics.{~Appendix~\ref{app:norm-dynamics} further confirms this at larger scale and relates the normalization to a time-varying effective momentum coefficient.}

% ============================================================
\subsection{Structural Decomposition for Orthogonalization Stability}
\label{sec:method-subspace}

After normalization, the momentum is ready for quantization. However, directly quantizing and then orthogonalizing it leads to severe error amplification. We now address this issue.

\begin{figure}[tbp]
    \centering
    \begin{subfigure}[b]{0.24\textwidth}
        \includegraphics[width=\textwidth]{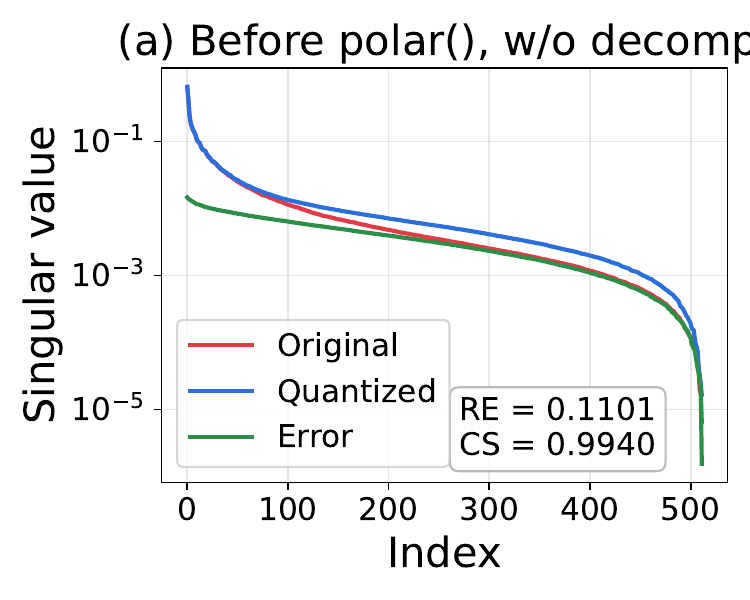}
    \end{subfigure}
    \hfill
    \begin{subfigure}[b]{0.24\textwidth}
        \includegraphics[width=\textwidth]{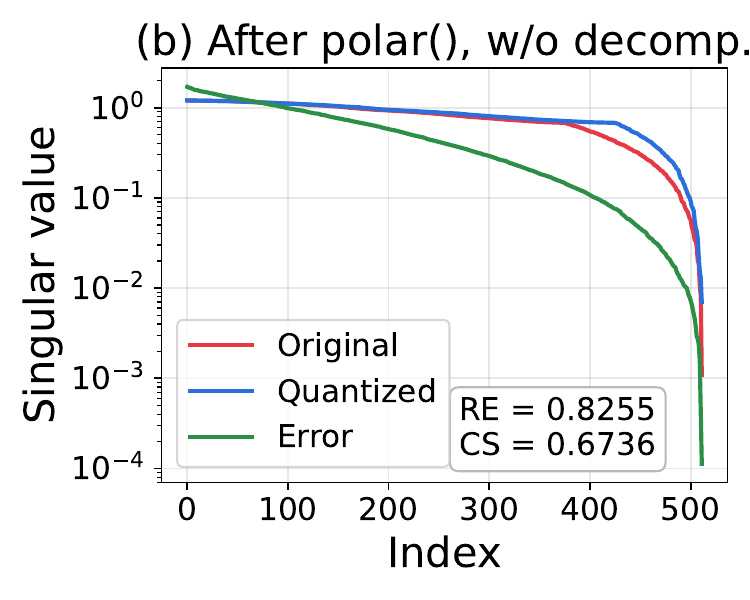}
    \end{subfigure}
    \hfill
    \begin{subfigure}[b]{0.24\textwidth}
        \includegraphics[width=\textwidth]{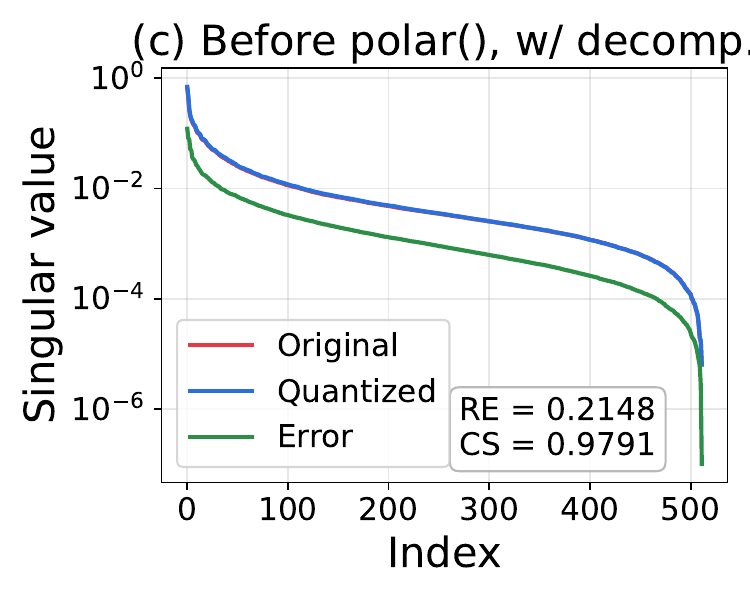}
    \end{subfigure}
    \hfill
    \begin{subfigure}[b]{0.24\textwidth}
        \includegraphics[width=\textwidth]{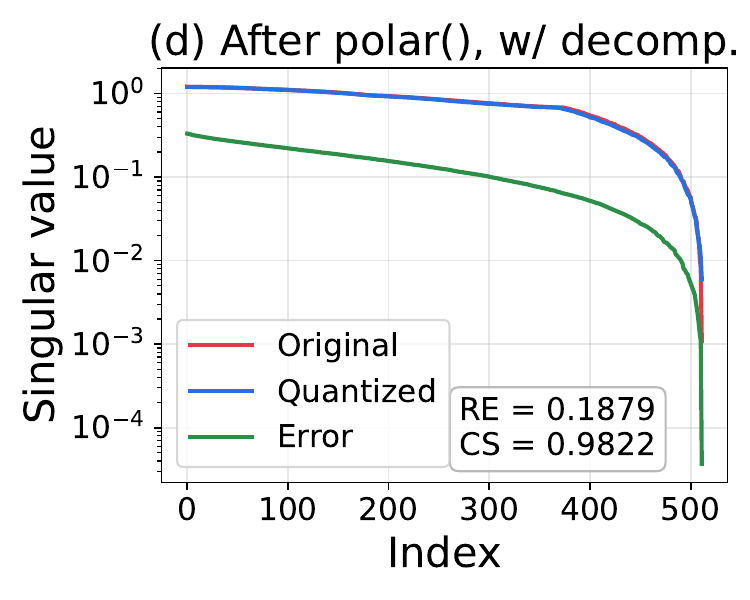}
    \end{subfigure}
    \vspace{-5pt}
    \caption{Singular value spectra of original and quantized momentum before and after orthogonalization on \texttt{layers.6.self\_attn.k\_proj}, with and without structural decomposition. Without decomposition (a, b), orthogonalization amplifies quantization error severely. With decomposition (c, d), directional fidelity is preserved through orthogonalization.}
    \label{fig:decompose}
\end{figure}

\paragraph{Error amplification through orthogonalization.}
The orthogonalization step amplifies the errors of the quantized momentum $\hat{\mat{M}}_t$  substantially~\citep{wu2026lowbit}, as shown in Figure~\ref{fig:decompose}(a, b).
This amplification has a clear spectral explanation. For $\mat{M}_t = \mat{U} \mat{\Sigma} \mat{V}^\top$, quantization perturbs both the singular values and the singular vectors. The polar decomposition maps all singular values to unity, which has two consequences: perturbations to singular value magnitudes are absorbed and vanish, but perturbations that rotate singular vector directions are fully retained and amplified, as they are no longer diluted by magnitude variation.

\paragraph{Structural decomposition.}
Our key insight is that if quantization errors only perturb the singular values without rotating the singular vectors, then the polar decomposition will naturally eliminate these errors. To achieve this, we decompose the momentum via SVD and quantize each factor independently:
\begin{equation}
    \mat{M}_t = \mat{U} \cdot \mat{S}, \quad \mat{S} = \mat{\Sigma} \mat{V}^\top,
\end{equation}
where $\mat{U} \in \mathbb{R}^{m \times r}$ contains the left singular vectors and $\mat{S} \in \mathbb{R}^{r \times n}$ encodes the singular values and right singular vectors, with $r = \min(m, n)$. The two singular factors are then quantized and dequantized separately:
\begin{equation}
    \hat{\mat{M}}_t = \mathrm{Dequant}(\mathrm{Quant}^{\mathrm{col}}_b(\mat{U})) \cdot \mathrm{Dequant}(\mathrm{Quant}^{\mathrm{row}}_b(\mat{S})).
\end{equation}

\begin{wrapfigure}{r}{0.45\textwidth}
    \centering
    \includegraphics[width=0.45\textwidth]{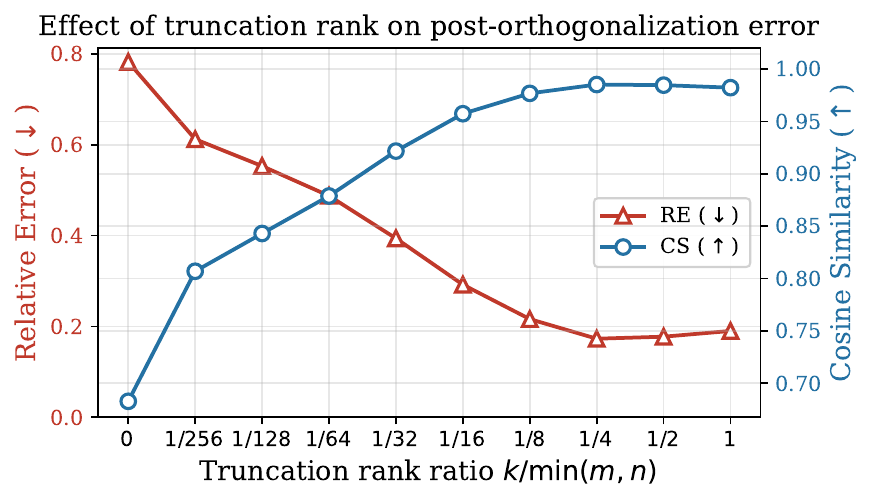}
    \caption{Effect of truncation rank $k$ on post-orthogonalization error under 4-bit companding quantization.}
    \vspace{-10pt}
    \label{fig:rank-ablation}
\end{wrapfigure}
Crucially, the quantization granularity is aligned with the singular structure: $\mat{U}$ is quantized \textbf{column-wise} so that each left singular vector $\mathbf{u}_i$ is quantized independently, and $\mat{S}$ is quantized \textbf{row-wise} so that each scaled right singular vector $\sigma_i \mathbf{v}_i^\top$ forms its own quantization group. Since the quantization scale factor is computed independently within each group, errors are confined to scaling each singular component individually without mixing or rotating different singular directions. Figure~\ref{fig:decompose}(c, d) confirms this: although decomposition slightly increases pre-orthogonalization error, it eliminates the error amplification caused by orthogonalization.

\paragraph{Truncation rank as a practical trade-off.}
In practice, polar approximation methods such as Newton--Schulz do not effectively amplify directions associated with very small singular values. {This follows from the construction of the iteration: a finite-step Newton--Schulz cannot push small singular values to unity, so error amplification concentrates in the dominant subspace and the top-$k$ truncation below is a meaningful trade-off.} Instead, error amplification is dominated by the principal singular subspace. Since quantization errors are primarily induced along the dominant singular directions of the original momentum~\citep{wu2026lowbit}, it suffices to align the quantization error with this subspace, without explicitly preserving the full spectrum. Concretely, we decompose:
\begin{equation}
    \mat{M}_t \approx \mat{U}_k \mat{S}_k + \mat{R}_k, \quad \mat{S}_k = \mat{\Sigma}_k \mat{V}_k^\top,
\end{equation}
where $\mat{U}_k \in \mathbb{R}^{m \times k}$ and $\mat{S}_k \in \mathbb{R}^{k \times n}$ capture the top-$k$ singular components, and $\mat{R}_k$ is the residual. Because the additional storage scales with $k$ rather than $\min(m,n)$, the memory overhead remains modest.
Following prior work~\citep{ahn2025dion,wu2026lowbit}, we adopt power iteration to extract the top-$k$ components efficiently. Given $\mat{S}_{t-1} \in \mathbb{R}^{k \times n}$ warm-started from the previous step, we iterate and recover the decomposition as:
\begin{equation}
    \mat{V}_{t-1} = \mathrm{RowNorm}(\mat{S}_{t-1}), \quad
    \mat{U}_t = \mathrm{orth}(\mat{M}_t \, \mat{V}_{t-1}^\top), \quad \mat{S}_t = \mat{U}_t^\top \mat{M}_t.
\end{equation}
Figure~\ref{fig:rank-ablation} shows the effect of the truncation rank on post-orthogonalization error. Both RE and CS improve rapidly with increasing $k$ and plateau around $k/\min(m,n) = 1/4$, indicating that a modest rank fraction is sufficient to capture the dominant directional error.

% ============================================================
\subsection{Companding Quantization for Dense-Region Distinguishability}
\label{sec:method-companding}
The final component of MuonQ optimizes the quantization operation itself. After normalization and decomposition, each component needs to be quantized to low precision. We now address how to design the quantizer for Muon's specific value distribution.
\paragraph{From outlier preservation to dense-region distinguishability.}
\begin{wrapfigure}{r}{0.45\textwidth}
    \centering
    \includegraphics[width=0.45\textwidth]{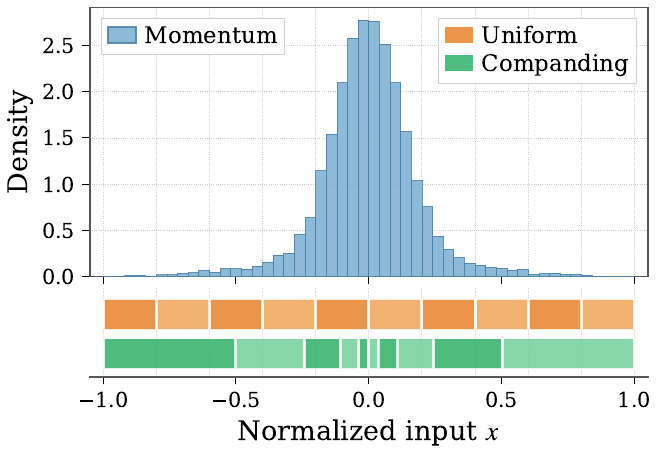}
    \vspace{-10pt}
    \caption{Normalized Muon momentum distribution (top) and quantization interval comparison (bottom). Each colored block represents one quantization bin.}
    \label{fig:companding}
    \vspace{-10pt}
\end{wrapfigure}
As illustrated in Figure~\ref{fig:companding}, the value distribution of $\mat{M}_t$ exhibits a sharp peak near zero with heavy tails. Under uniform quantization, equal-width bins waste resolution on sparsely populated tails while providing insufficient precision for the dense center. For Adam, its element-wise update is dominated by large-magnitude states, so prior quantization methods~\citep{dettmers2022optimizers} need only preserve outliers and can safely ignore the dense region. However, Muon's polar decomposition equalizes all singular directions, meaning the dense, small-magnitude entries carry equal directional importance. The critical requirement thus shifts from outlier preservation to \textbf{distinguishability} among closely packed values near zero.
\paragraph{$\mu$-law companding quantization.}
We address this resolution imbalance by applying companding quantization, a classical technique from signal processing~\citep{smith1957companding} that reshapes the value distribution prior to quantization. 
Companding wraps the base quantizer with a nonlinear transform: a compressing function $f$ is applied before quantization, and its inverse $f^{-1}$ after dequantization. We adopt the $\mu$-law function~\citep{mulaw}:
\begin{equation}
    f(x) = \mathrm{sgn}(x)\,\frac{\ln(1 + \mu\,|x|)}{\ln(1 + \mu)}, \quad -1 \leq x \leq 1,
\label{eq:mulaw}
\end{equation}
where we set $\mu = 255$ following the standard $\mu$-law convention in digital telephony~\citep{mulaw}. 
The companded quantization and dequantization operators are defined as:
\begin{equation}
    \label{eq:companding-quant}
    \mat{q}, \mat{s} = \mathrm{CQuant}_b^{g}(\mat{x}) = \mathrm{Quant}_b^{g}(f(\mat{x})), \quad
    \hat{\mat{x}} = \mathrm{CDequant}(\mat{q}, \mat{s}) = f^{-1}(\mathrm{Dequant}(\mat{q}, \mat{s})).
\end{equation}

\begin{wraptable}{r}{0.5\textwidth}
    \centering
    \vspace{-5pt}
    \small
    \setlength{\tabcolsep}{3.5pt}
    \begin{tabular}{lcccc}
        \toprule
        \multirow{2}{*}{Gran.} & \multicolumn{2}{c}{RE ($\downarrow$)} & \multicolumn{2}{c}{CS ($\uparrow$)} \\
        \cmidrule(lr){2-3} \cmidrule(lr){4-5}
        & Uni. & Comp. & Uni. & Comp. \\
        \midrule
        Tensor & 0.509 & \textbf{0.238} {\color{green!60!black}\scriptsize($\downarrow$.271)} & 0.879 & \textbf{0.973} {\color{green!60!black}\scriptsize($\uparrow$.094)} \\
        Row    & 0.127 & \textbf{0.111} {\color{green!60!black}\scriptsize($\downarrow$.016)} & 0.992 & \textbf{0.994} {\color{green!60!black}\scriptsize($\uparrow$.002)} \\
        Column & 0.254 & \textbf{0.159} {\color{green!60!black}\scriptsize($\downarrow$.095)} & 0.969 & \textbf{0.988} {\color{green!60!black}\scriptsize($\uparrow$.019)} \\
        \bottomrule
    \end{tabular}
    \caption{Effect of $\mu$-law companding on 4-bit quantization of \texttt{layers.0.self\_attn.k\_proj}. }
    \label{tab:companding}
    \vspace{-5pt}
\end{wraptable}
As illustrated in Figure~\ref{fig:companding}, the logarithmic nonlinearity of $f$ reshapes how quantization bins are allocated: the {\color{orange}uniform} band divides the input range into equal-width bins, whereas the {\color{green!60!black}companding} band concentrates narrower bins near zero while assigning wider bins to the sparsely populated tails. As shown in Table~\ref{tab:companding}, this density-aware reallocation consistently improves reconstruction quality across all granularities.

\begin{figure}[t]
    \centering
    \vspace{-10pt}
    \begin{subfigure}[b]{0.32\linewidth}
        \caption{GPT-2 Medium}
        \includegraphics[width=\linewidth]{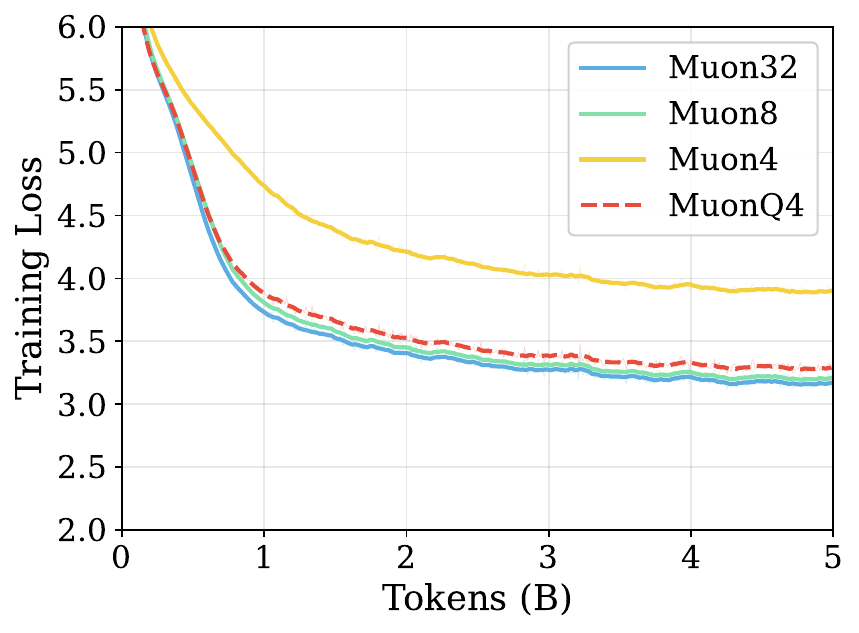}
    \end{subfigure}
    \hfill
    \begin{subfigure}[b]{0.32\linewidth}
        \caption{LLaMA-350M}
        \includegraphics[width=\linewidth]{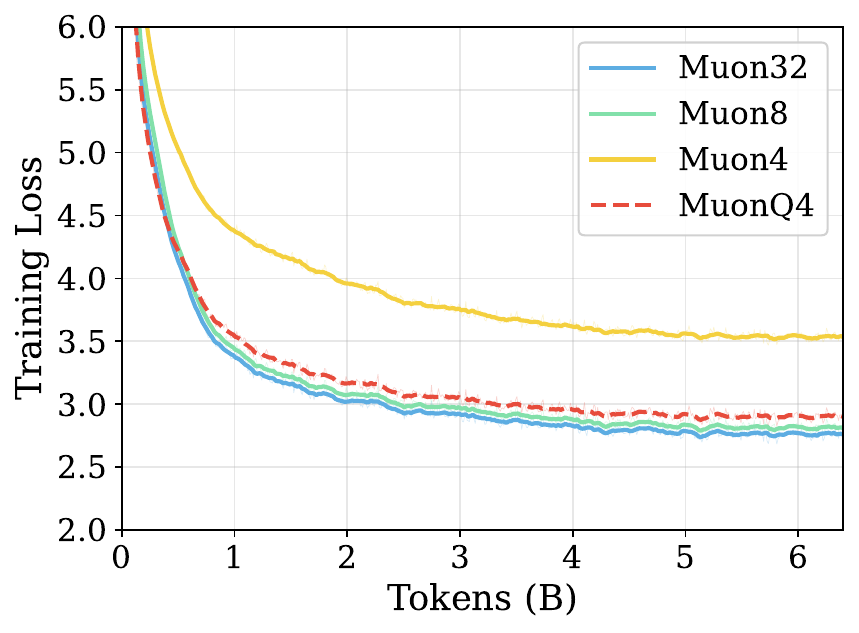}
    \end{subfigure}
    \hfill
    \begin{subfigure}[b]{0.32\linewidth}
        \caption{LLaMA-1.1B}
        \includegraphics[width=\linewidth]{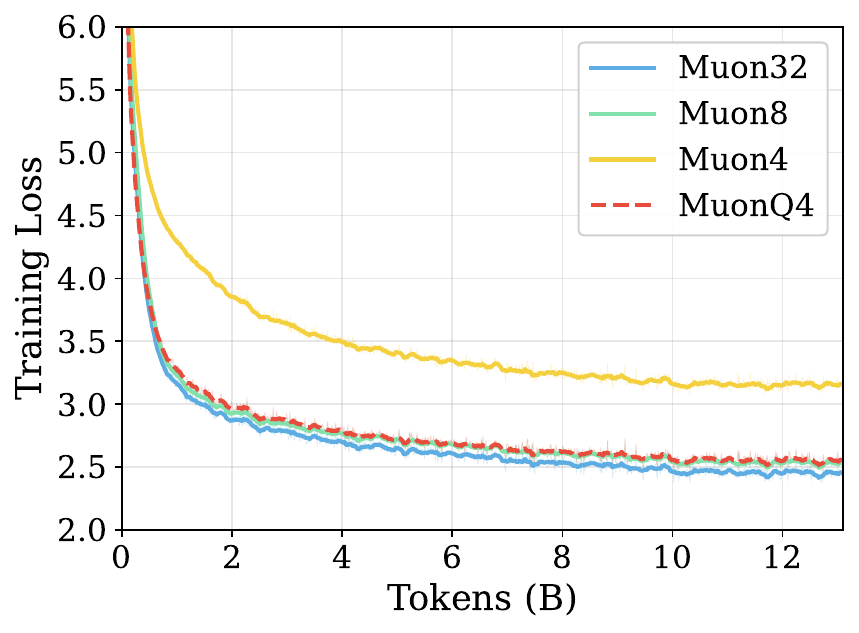}
    \end{subfigure}
    \caption{Training loss curves for GPT-2 and LLaMA models on FineWeb. {MuonQ4 closely tracks full-precision Muon32 throughout training, while naive Muon4 retains a persistent loss gap that grows with model scale.}}
    \label{fig:training-curves}
\end{figure}
\section{Experiments}

We evaluate MuonQ on two model families: GPT-2 (Medium, Large) and LLaMA (350M, 1.1B), all trained on FineWeb~\citep{fineweb} following the standard Muon protocol~\citep{liu2025muon}. We compare against three baselines: full-precision Muon (Muon32), and uniform quantization at 8-bit (Muon8) and 4-bit (Muon4){; we additionally compare against the mixed-precision method GRASP~\citep{wu2026lowbit} (\S\ref{sec:main-results})}. All methods share identical training hyperparameters and differ only in momentum representation. Unless otherwise noted, we apply tensor-wise quantization to the full momentum matrix $\mat{M}_t$ (Muon8/Muon4) or the residual $\mat{R}_t$ (MuonQ4) and a truncation rank of $k = 1/16$. Detailed configurations are provided in Appendix~\ref{app:training-details}. All experiments are conducted on NVIDIA A100 GPUs.

% We evaluate MuonQ on two model families, GPT-style and LLaMA-style models, all trained on FineWeb~\citep{fineweb}. Across all experiments, we follow the standard Muon training protocol~\citep{liu2025muon}: Muon is applied to matrix-shaped hidden-layer parameters, while AdamW is used for embeddings and the language modeling head. Detailed model architectures and training hyperparameters are provided in Appendix~\ref{app:training-details}.

% We compare MuonQ against Muon with full-precision momentum (Muon32) as the baseline, as well as naive uniform quantization at 8-bit (Muon8) and 4-bit (Muon4) precision. Unless otherwise specified, all methods share identical training hyperparameters and differ only in the representation of the momentum.
% Except for the ablation studies, all experiments adopt tensor-wise quantization, where each momentum tensor is associated with a single scale factor.
% For MuonQ, we fix the truncation rank to $1/16$ of the original by default, which provides a good trade-off between accuracy and efficiency. All experiments are conducted on NVIDIA A100 GPUs.

\subsection{Main Results}
\label{sec:main-results}
\paragraph{Training Dynamics.}
Figure~\ref{fig:training-curves} shows the training loss curves across representative model configurations throughout training. Across all scales, MuonQ4 closely tracks the full-precision Muon32 baseline, demonstrating stable convergence from the early stages of training through to the end. In contrast, naive Muon4 exhibits a persistent loss gap that does not diminish over the course of training. Notably, the advantage of MuonQ4 over naive quantization becomes more pronounced as model scale increases, suggesting that preserving directional fidelity plays an increasingly critical role at larger scales.

\paragraph{Downstream evaluation.}
Table~\ref{tab:main-downstream} reports zero-shot accuracy on ARC-Challenge, ARC-Easy~\citep{clark2018arc}, OpenBookQA~\citep{mihaylov2018openbookqa}, BoolQ~\citep{clark2019boolq}, HellaSwag~\citep{zellers2019hellaswag}, PIQA~\citep{bisk2020piqa}, and WinoGrande~\citep{sakaguchi2021winogrande} across all model scales and architectures, evaluated using the \texttt{lm-evaluation-harness} library~\citep{gao2023lmeval}. MuonQ4 consistently matches the Muon32 baseline on various benchmarks, demonstrating that our quantization preserves downstream task performance.

\begin{table}[tbp]
    \centering
    \small
    \setlength{\tabcolsep}{5pt}
    \begin{tabular}{cc cccccccc}
        \toprule
        \textbf{Model} & \textbf{Optimizer} & \textbf{ARC-c} & \textbf{ARC-e} & \textbf{OBQA} & \textbf{BoolQ} & \textbf{HellaS.} & \textbf{PIQA} & \textbf{WinoG.} & \textbf{Avg.} \\
        \midrule
        \multirow{4}{*}{\shortstack{\textbf{GPT-2} \\ \textbf{Medium}}}
        & Muon32  & 23.5 & 39.7 & 27.8 & 57.4 & 33.1 & 65.1 & 51.3 & 42.6 \\
        & Muon8   & 24.4 & 39.5 & 29.0 & 56.4 & 32.1 & 64.6 & 51.2 & 42.5 \\
        & \cellcolor{gray!25} Muon4   & \cellcolor{gray!25} 22.4 & \cellcolor{gray!25} 31.9 & \cellcolor{gray!25} 25.0 & \cellcolor{gray!25} 62.0 & \cellcolor{gray!25} 26.6 & \cellcolor{gray!25} 58.2 & \cellcolor{gray!25} 50.6 & \cellcolor{gray!25} 39.5 \\
        & \cellcolor{gray!25} \textbf{MuonQ4} & \cellcolor{gray!25} 23.9 & \cellcolor{gray!25} 37.8 & \cellcolor{gray!25} 25.6 & \cellcolor{gray!25} 60.7 & \cellcolor{gray!25} 30.2 & \cellcolor{gray!25} 63.1 & \cellcolor{gray!25} 50.8 & \cellcolor{gray!25} \textbf{41.7} \\
        \cmidrule(lr){1-10}
        \multirow{4}{*}{\shortstack{\textbf{GPT-2} \\ \textbf{Large}}}
        & Muon32  & 24.0 & 42.6 & 29.4 & 56.4 & 37.9 & 67.2 & 49.4 & 43.8 \\
        & Muon8   & 23.7 & 41.0 & 30.0 & 59.8 & 36.2 & 66.7 & 50.1 & 43.9 \\
        & \cellcolor{gray!25} Muon4   & \cellcolor{gray!25} 21.1 & \cellcolor{gray!25} 35.1 & \cellcolor{gray!25} 23.4 & \cellcolor{gray!25} 61.8 & \cellcolor{gray!25} 27.5 & \cellcolor{gray!25} 58.2 & \cellcolor{gray!25} 50.0 & \cellcolor{gray!25} 39.6 \\
        & \cellcolor{gray!25} \textbf{MuonQ4} & \cellcolor{gray!25} 22.8 & \cellcolor{gray!25} 39.8 & \cellcolor{gray!25} 30.2 & \cellcolor{gray!25} 59.8 & \cellcolor{gray!25} 33.8 & \cellcolor{gray!25} 65.2 & \cellcolor{gray!25} 51.5 & \cellcolor{gray!25} \textbf{43.3} \\
        \cmidrule(lr){1-10}
        \multirow{4}{*}{\shortstack{\textbf{LLaMA} \\ \textbf{350M}}}
        & Muon32  & 22.6 & 38.5 & 28.4 & 62.0 & 34.3 & 65.7 & 51.9 & 43.3 \\
        & Muon8   & 22.9 & 38.2 & 27.6 & 61.2 & 32.3 & 63.4 & 53.7 & 42.8 \\
        & \cellcolor{gray!25} Muon4   & \cellcolor{gray!25} 21.5 & \cellcolor{gray!25} 29.6 & \cellcolor{gray!25} 25.0 & \cellcolor{gray!25} 61.9 & \cellcolor{gray!25} 27.0 & \cellcolor{gray!25} 55.9 & \cellcolor{gray!25} 50.0 & \cellcolor{gray!25} 38.7 \\
        & \cellcolor{gray!25} \textbf{MuonQ4} & \cellcolor{gray!25} 22.4 & \cellcolor{gray!25} 38.1 & \cellcolor{gray!25} 27.8 & \cellcolor{gray!25} 60.9 & \cellcolor{gray!25} 31.0 & \cellcolor{gray!25} 63.7 & \cellcolor{gray!25} 51.6 & \cellcolor{gray!25} \textbf{42.2} \\
        \cmidrule(lr){1-10}
        \multirow{4}{*}{\shortstack{\textbf{LLaMA} \\ \textbf{1.1B}}}
        & Muon32  & 26.4 & 45.2 & 30.4 & 60.9 & 45.9 & 69.6 & 51.3 & 47.1 \\
        & Muon8   & 24.4 & 42.3 & 31.4 & 61.3 & 41.2 & 69.6 & 52.2 & 46.1 \\
        & \cellcolor{gray!25} Muon4   & \cellcolor{gray!25} 22.2 & \cellcolor{gray!25} 34.1 & \cellcolor{gray!25} 25.6 & \cellcolor{gray!25} 60.3 & \cellcolor{gray!25} 28.5 & \cellcolor{gray!25} 58.7 & \cellcolor{gray!25} 49.6 & \cellcolor{gray!25} 39.8 \\
        & \cellcolor{gray!25} \textbf{MuonQ4} & \cellcolor{gray!25} 25.0 & \cellcolor{gray!25} 41.8 & \cellcolor{gray!25} 30.4 & \cellcolor{gray!25} 60.7 & \cellcolor{gray!25} 40.3 & \cellcolor{gray!25} 68.3 & \cellcolor{gray!25} 49.9 & \cellcolor{gray!25} \textbf{45.2} \\
        \bottomrule
    \end{tabular}
    \caption{Zero-shot accuracy (\%, $\uparrow$) on downstream benchmarks. {MuonQ4 trails Muon32 by only about one point on average and recovers most of the accuracy lost by naive Muon4.}}
    \label{tab:main-downstream}
\end{table}

\begin{figure}[tbp]
    \centering
    \includegraphics[width=\textwidth]{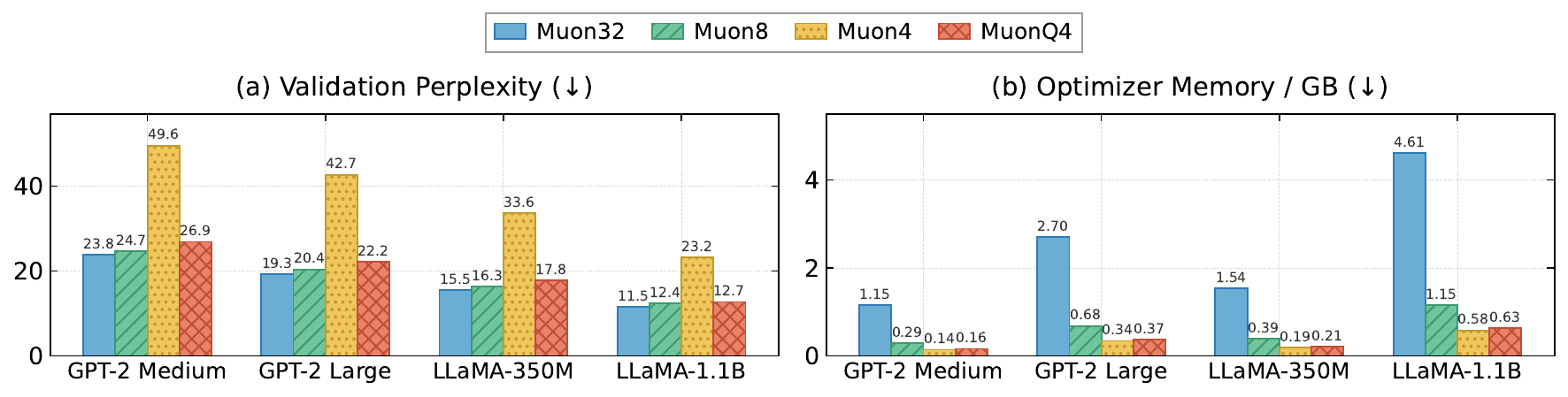}
    \caption{Validation PPL ($\downarrow$) and optimizer-state memory (GB) across model scales. {MuonQ4 substantially narrows the PPL gap to Muon32 while achieving up to $7.3\times$ optimizer-state memory reduction.}}
    \label{fig:efficiency}
\end{figure}

\paragraph{Memory efficiency comparison.}
Figure~\ref{fig:efficiency} compares validation perplexity and optimizer-state memory across model scales, relative to the naive baselines. {Across all scales MuonQ4 substantially narrows the PPL gap to the Muon32 baseline while reducing optimizer memory by $7.3\times$ on average.} Compared to Muon4, MuonQ4 incurs only a modest memory increase but recovers the majority of the full-precision performance.

% >>> CR-NEW (camera-ready, Reviewer gTtK W1/W2 + Reviewer CBUN): GRASP head-to-head + wall-clock
{\paragraph{Comparison with GRASP and training efficiency.}
We compare against the mixed-precision method GRASP~\citep{wu2026lowbit} on LLaMA-1.1B under settings identical to our main results, reporting per-step wall-clock time on the same hardware (Table~\ref{tab:grasp-headtohead}). MuonQ4 delivers the highest compression of any method we compare, reducing optimizer-state memory to $0.62$\,GB ($7.28\times$) against GRASP's $0.75$\,GB ($6.02\times$), while matching the 8-bit and mixed-precision baselines in perplexity to within $0.3$ ($12.67$ vs.\ Muon8's $12.42$ and GRASP's $12.33$). Pure 4-bit MuonQ4 thus reaches the best memory efficiency at essentially no accuracy cost relative to these stronger baselines. Because MuonQ modifies only the optimizer step, its per-step overhead is $+10.72\%$ over Muon32, on par with GRASP's $+9.64\%$. This is an artifact of our large-batch setup, where the optimizer occupies an unusually large share of per-step time; at production scale this share shrinks, keeping the overhead well within the $\sim\!2\times$ speedup Muon provides over AdamW.}

% >>> CR-NEW (camera-ready): Table for the GRASP head-to-head + wall-clock (LLaMA-1.1B)
\begin{table}[tbp]
    \centering
    \small
    \setlength{\tabcolsep}{6pt}
    \begin{tabular}{lccccc}
        \toprule
        \textbf{Method} & \textbf{Val PPL}$\downarrow$ & \textbf{Opt.\ Mem.} & \textbf{Mem.}$\downarrow$ & \textbf{Step Time} & \textbf{Overhead} \\
        \midrule
        Muon32          & 11.47 & 4.50\,GB & 1.00$\times$ & 4408.5\,ms & --- \\
        Muon8           & 12.42 & 1.13\,GB & 4.00$\times$ & 4433.8\,ms & +0.58\% \\
        Muon4           & 23.18 & 0.56\,GB & 8.00$\times$ & 4482.3\,ms & +1.68\% \\
        GRASP (8/4-bit) & \textbf{12.33} & 0.75\,GB & 6.02$\times$ & 4833.4\,ms & +9.64\% \\
        \rowcolor{gray!25} \textbf{MuonQ4 (ours)} & 12.67 & \textbf{0.62\,GB} & \textbf{7.28$\times$} & 4881.2\,ms & +10.72\% \\
        \bottomrule
    \end{tabular}
    \caption{{Head-to-head comparison against GRASP~\citep{wu2026lowbit} on LLaMA-1.1B  under identical training settings; per-step time is measured on the same hardware. MuonQ4 attains the largest memory reduction, trading a small PPL increase for additional memory savings rather than dominating these baselines.}}
    \label{tab:grasp-headtohead}
\end{table}

\subsection{Ablation Study}
\label{sec:ablation}

In this section, we ablate the design choices of MuonQ on GPT-2 Small (124M) trained with 1B FineWeb tokens. The best configuration identified here is used for main results in \S\ref{sec:main-results}.

\paragraph{Component ablation.}
\begin{wraptable}{r}{0.55\textwidth}
    \centering
    \vspace{-10pt}
    \small
    \begin{tabular}{lccc cc}
        \toprule
        \textbf{Method} & \textbf{C} & \textbf{N} & \textbf{D} & \textbf{PPL}$\downarrow$ & \textbf{Mem.}$\downarrow$ \\
        \midrule
        Muon32 & --         & --         & --         & 36.4 & 324.0 \\
        Muon4  & --         & --         & --         & 69.9 & 40.5 {\color{red}\scriptsize(8.0$\times$)} \\
        \midrule
        \textbf{MuonQ4} & \checkmark & \ding{55}  & \ding{55}  & 66.6 {\color{green!60!black}\scriptsize($\downarrow$3.3)} & 40.5 {\color{red}\scriptsize(8.0$\times$)} \\
        \textbf{MuonQ4} & \ding{55}  & \checkmark & \ding{55}  & 50.0 {\color{green!60!black}\scriptsize($\downarrow$19.9)} & 40.5 {\color{red}\scriptsize(8.0$\times$)} \\
        \textbf{MuonQ4} & \checkmark & \checkmark & \ding{55}  & 46.2 {\color{green!60!black}\scriptsize($\downarrow$23.7)} & 40.5 {\color{red}\scriptsize(8.0$\times$)} \\
        \textbf{MuonQ4} & \ding{55}  & \ding{55}  & \checkmark & 44.7 {\color{green!60!black}\scriptsize($\downarrow$25.2)} & 44.3 {\color{red}\scriptsize(7.3$\times$)} \\
        \midrule
        \textbf{MuonQ4} & \checkmark & \checkmark & \checkmark & \textbf{40.9} {\color{green!60!black}\scriptsize($\downarrow$29.0)} & 44.3 {\color{red}\scriptsize(7.3$\times$)} \\
        \bottomrule
    \end{tabular}
    \caption{Component ablation on GPT-2 Small. \textbf{C}: companding, \textbf{N}: normalization, \textbf{D}: decomposition.}
    \label{tab:component-ablation}
    \vspace{-20pt}
\end{wraptable}
Table~\ref{tab:component-ablation} isolates the effect of each technique. Normalization and companding each improve PPL independently with negligible memory overhead, and their combination yields further gains. Structural decomposition introduces a modest memory increase due to the decomposed factors but provides the largest PPL improvement by protecting singular directions through orthogonalization. The full MuonQ achieves the best trade-off between PPL and memory.

\paragraph{Truncation rank ratio selection.}
\begin{wrapfigure}{r}{0.45\textwidth}
    \centering
    \vspace{-10pt}
    % 占位：rank vs PPL/Memory/Time 的 tradeoff 图
    \includegraphics[width=0.45\textwidth]{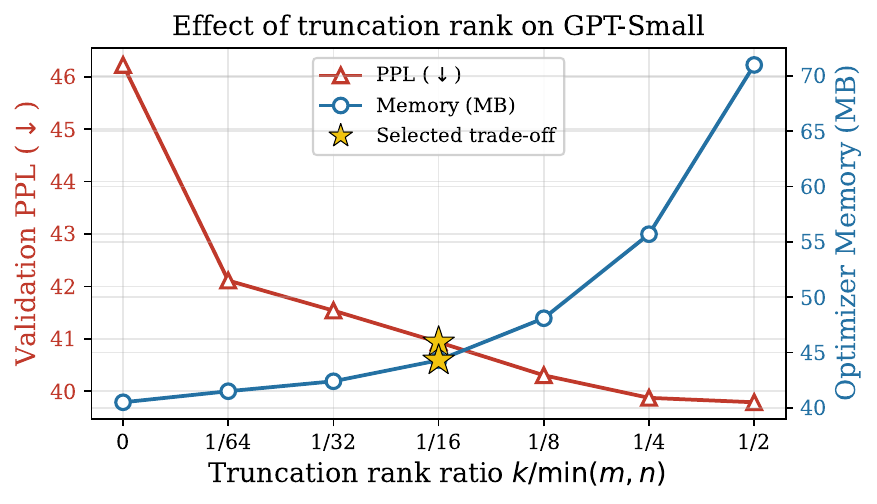}
    \caption{Effect of truncation rank on end-to-end training. Left: validation PPL ($\downarrow$). Right: optimizer state memory (MB).}
    \label{fig:rank-ablation-e2e}
    \vspace{-10pt}
\end{wrapfigure}
Figure~\ref{fig:rank-ablation-e2e} shows the trade-off between validation PPL and optimizer memory as the truncation rank $k$ varies. Increasing $k$ monotonically improves PPL by preserving more singular directions, but the memory overhead grows due to the additional storage of $U_t$ and $S_t$. PPL improves sharply from $k=0$ (no decomposition) to $k=\min(m,n)/16$, after which the gains diminish while memory continues to rise. We select $k=\min(m,n)/16$ as the default, as it achieves most of the PPL improvement with only a 9.4\% memory increase over the baseline.

\begin{figure}[tbp]
    \centering
    \includegraphics[width=\textwidth]{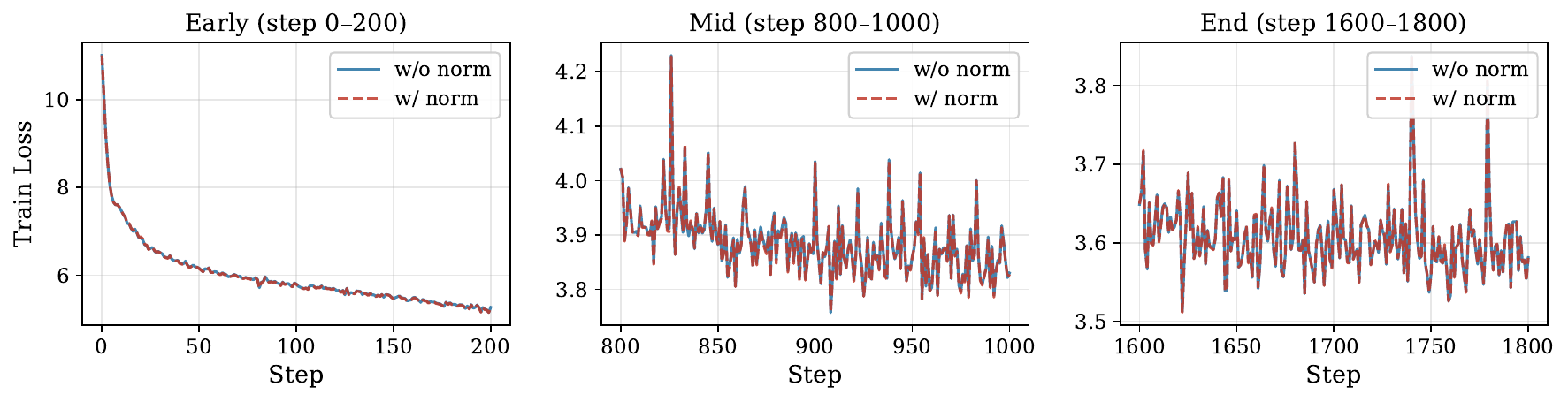}
    \caption{Training loss of full-precision Muon with and without pre-quantization normalization on GPT-2 Small, shown at early, mid, and late stages.}
    \label{fig:norm-ablation}
    \vspace{-10pt}
\end{figure}
\paragraph{Normalization without quantization}
\label{app:norm-without-quant}

Our pre-quantization normalization (\S\ref{sec:method-normalize}) modifies the momentum recursion by rescaling both the gradient and the momentum at each step. While this changes the relative magnitudes of successive momentum updates, it does not affect the parameter update direction.  Figure~\ref{fig:norm-ablation} shows the training loss at three stages of training. The two curves are virtually indistinguishable throughout, with a mean absolute difference of only 0.003 in training loss, confirming that the normalization introduces no measurable perturbation to the optimization dynamics under full precision.

\paragraph{Quantization granularity.}
\begin{wraptable}{r}{0.38\textwidth}
    \centering
    \vspace{-10pt}
    \setlength{\tabcolsep}{4pt}
    \small
    \begin{tabular}{l cc}
        \toprule
        \multirow{2}{*}{\textbf{Gran.}} & \multicolumn{2}{c}{\textbf{PPL} $\downarrow$} \\
        \cmidrule(lr){2-3}
        & Muon4 & \textbf{MuonQ4} \\
        \midrule
        \textbf{Tensor} & 69.9 & 40.9 {\color{green!60!black}\scriptsize($\downarrow$29.0)} \\
        \textbf{Column} & 43.4 & 39.6 {\color{green!60!black}\scriptsize($\downarrow$3.8)} \\
        \textbf{Row}    & 42.9 & 39.5 {\color{green!60!black}\scriptsize($\downarrow$3.4)} \\
        \bottomrule
    \end{tabular}
    \caption{Granularity ablation.}
    \vspace{-10pt}
    \label{tab:gran-ablation}
    
\end{wraptable}

Table~\ref{tab:gran-ablation} compares quantization granularities across methods. MuonQ4 consistently outperforms the Muon4 baseline at every granularity level. Notably, even MuonQ4 with the coarsest tensor-wise granularity (PPL 40.9) surpasses the best Muon4 configuration with row-wise quantization (PPL 42.9), demonstrating that our techniques provide improvements orthogonal to the choice of granularity.

\section{Related Works}
\label{sec:related works}
\paragraph{Memory-Efficient Optimizers} 
The memory overhead of optimizer states has motivated a rich line of research on efficient alternatives. One family of approaches exploits low-rank or factored structure to reduce state size. Adafactor~\citep{shazeer2018adafactor} and CAME~\citep{luo2023came} approximate Adam's second moment via factored representations, while GaLore~\citep{zhao2024galore} and Flora~\citep{hao2024flora} project gradients into low-rank subspaces before applying the optimizer. {COSMOS~\citep{liu2025cosmos} similarly exploits the leading spectral subspace, using an SVD-based projection to build a memory-efficient hybrid preconditioner.} Sign-based methods such as SignSGD~\citep{bernstein2018signsgd} and 1-bit Adam~\citep{tang2021onebit} compress gradients or momentum to their signs, reducing both memory and communication costs. Other approaches simplify the optimizer state itself: SM3~\citep{anil2019sm3} maintains memory proportional to the sum rather than the product of parameter dimensions, and LAMB~\citep{you2020lamb} replaces per-element states with layerwise adaptive rates. LoRA~\citep{hu2022lora} takes a different approach by freezing pretrained weights and tuning only low-rank adapters.

\paragraph{Quantization of Optimizer States}

A parallel line of work compresses optimizer states via low-precision representation.
For Adam and its variants, \citet{dettmers2022optimizers} proposed block-wise dynamic quantization that enables stable 8-bit optimizer states, and subsequent work has pushed compression to 4-bit for both first-order moments~\citep{li2023fourbit} and second-order preconditioners~\citep{wang2024fourbit,zhang2024qgalore}.
For the Muon optimizer, \citet{gupta2025quantmuon} showed that 8-bit blockwise quantization preserves performance under both linear and dynamic schemes.
\citet{wu2026lowbit} identified that orthogonalization amplifies quantization error primarily in the top singular subspace, and proposed preserving these components at 8-bit while compressing the residual to 4-bit with grid quantization.
Theoretical analysis has further suggested that Muon is inherently more robust to quantization than Adam~\citep{tang2026convergence}.

\section{Conclusion}

We have presented MuonQ, a framework for low-bit quantization of Muon's optimizer states guided by the principle of directional fidelity. Three complementary techniques address quantization errors at different stages of the Muon pipeline: pre-quantization normalization stabilizes per-step error magnitudes to prevent directional drift accumulation; structural decomposition via power iteration ensure quantization errors perturb only singular value without rotating singular vectors; and $\mu$-law companding reallocates quantization bins toward the dense center of the momentum distribution where distinguishability matters most. Together, these enable stable 4-bit quantization with up to 7.3$\times$ optimizer memory reduction. {Experiments across GPT-2 and LLaMA models demonstrate that MuonQ recovers most of full-precision Muon's training loss and downstream accuracy at a favorable point on the accuracy--memory trade-off.}

\paragraph{Limitations and future work.}
Our experiments are conducted at moderate model scales (up to 1.1B parameters), and the structural decomposition introduces per-step overhead from power iteration. Scaling MuonQ to frontier-scale training with distributed parallelism, combining it with gradient compression techniques such as GaLore~\citep{zhao2024galore} and low-rank adaptation~\citep{hu2022lora}, and extending to sub-4-bit or mixed-format (FP8/FP4) regimes are promising directions for future work.

\section{Acknowledgement}
This material is based upon work supported by the U.S. Department of Energy, Office of Science, Office of Advanced Scientific Computing Research, Artificial Intelligence for Science program, under contract DE-SC0025390. This research used resources of the National Energy Research Scientific Computing Center, a DOE Office of Science User Facility supported by the Office of Science of the U.S. Department of Energy under Contract No. DE-AC02-05CH11231 using NERSC award ASCR-ERCAP0030039, as well as NERSC award ALCCERCAP0031379.

\clearpage
\bibliography{colm2026_conference}
\bibliographystyle{colm2026_conference}

\newpage
\appendix
\section{Training Details}
\label{app:training-details}

\subsection{Model Architectures}

We evaluate on two model families. Table~\ref{tab:model-arch} summarizes the architecture configurations.

\begin{table}[htbp]
    \centering
    \small
    \begin{tabular}{ll cccccc}
        \toprule
        Family & Model & $d_{\text{model}}$ & $n_{\text{layer}}$ & $n_{\text{head}}$ & $d_{\text{ffn}}$ & Vocab & Context \\
        \midrule
        \multirow{3}{*}{GPT-2}
        & Small  & 768  & 12 & 12 & 3072  & 50257 & 1024 \\
        & Medium & 1024 & 24 & 16 & 4096  & 50257 & 4096 \\
        & Large  & 1280 & 36 & 20 & 5120  & 50257 & 8192 \\
        \midrule
        \multirow{2}{*}{LLaMA}
        & 350M & 1024 & 24 & 16 & 2736 & 32000 & 4096 \\
        & 1.1B & 2048 & 24 & 32 & 5461 & 32000 & 4096 \\
        \bottomrule
    \end{tabular}
    \caption{Model architecture configurations.}
    \label{tab:model-arch}
\end{table}

GPT-2 models use learned positional embeddings and the GPT-2 tokenizer (vocab size 50257). LLaMA models use RoPE positional encoding ($\theta = 10000$) and the LLaMA-2 tokenizer (vocab size 32000). All models use FlashAttention.

\subsection{Training Configuration}

Table~\ref{tab:training-config} summarizes the training configuration. All models are trained for a single epoch with BF16 mixed precision, gradient clipping at 1.0, and \texttt{torch.compile} enabled.

\begin{table}[htbp]
    \centering
    \small
    \begin{tabular}{ll ccccc}
        \toprule
        Family & Model & Tokens & GPUs & Batch/GPU & Tokens/Step & Dataset \\
        \midrule
        \multirow{3}{*}{GPT-2}
        & Small  & 1B   & 8$\times$A100  & 32 & 524K  & FineWeb \\
        & Medium & 5B   & 32$\times$A100 & 4  & 2.1M  & FineWeb \\
        & Large  & 10B  & 32$\times$A100 & 2  & 4.2M  & FineWeb \\
        \midrule
        \multirow{2}{*}{LLaMA}
        & 350M & 6.4B  & 32$\times$A100 & 4 & 2.1M & FineWeb \\
        & 1.1B & 13.1B & 32$\times$A100 & 4 & 2.1M & FineWeb \\
        \bottomrule
    \end{tabular}
    \caption{Training configurations. Batch/GPU denotes the per-device batch size. The global tokens per step is computed as GPUs $\times$ Batch/GPU $\times$ context length $\times$ gradient accumulation steps, where gradient accumulation is adjusted to match the target tokens/step.}
    \label{tab:training-config}
\end{table}

\subsection{Optimizer Configuration}

All experiments use the Muon optimizer for matrix-shaped hidden-layer parameters and AdamW for embeddings and the language modeling head, following the standard Muon training protocol~\citep{liu2025muon}. Table~\ref{tab:optim-config} summarizes the optimizer hyperparameters.

\begin{table}[htbp]
    \centering
    \small
    \begin{tabular}{lcccc}
        \toprule
        & Muon32 & Muon8 & Muon4 & MuonQ4 \\
        \midrule
        Learning rate       & 0.001 & 0.001 & 0.001 & 0.001 \\
        Weight decay        & 0.1   & 0.1   & 0.1   & 0.1   \\
        Polar method        & NS5   & NS5   & NS5   & NS5   \\
        Bit-width           & 16    & 8     & 4     & 4     \\
        Granularity         & --    & tensor & tensor & tensor \\
        Companding ($\mu$)  & --    & --    & --    & 255   \\
        Normalization       & --    & --    & --    & \checkmark \\
        Rank ($k$)          & --    & --    & --    & $\min(m,n)/16$ \\
        \bottomrule
    \end{tabular}
    \caption{Optimizer configurations. All variants share the same base hyperparameters; only the momentum quantization scheme differs.}
    \label{tab:optim-config}
\end{table}

AdamW uses a learning rate of 0.001 and weight decay of 0.1 for the embedding and LM head parameters across all experiments. NS5 denotes 5-step Newton--Schulz iteration with the standard coefficients $(a, b, c) = (3.4445, -4.7750, 2.0315)$~\citep{jordan2024muon}.

\section{Comparison with 4-bit-GRASP}
\label{app:grasp}

{This appendix isolates the effect of \emph{bit-width} by holding granularity fixed; an end-to-end against the officially released GRASP code, is reported in the main text (\S\ref{sec:main-results}, Table~\ref{tab:grasp-headtohead}).}

\paragraph{Methodological difference.}
4bit-Muon-GRASP~\citep{wu2026lowbit} shares with MuonQ the observation that orthogonalization amplifies quantization error primarily in the top singular subspace.
To address this, GRASP stores the top-$k$ singular components at 8-bit precision while compressing the residual to 4-bit, constituting a mixed-precision scheme.
However, GRASP does not analyze why decomposing the momentum is itself beneficial independent of the bit-width assigned to the factors.
As we showed in \S\ref{sec:method-subspace}, the key mechanism is \textbf{aligning quantization granularity with the singular structure}: column-wise quantization of $\mathbf{U}$ and row-wise quantization of $\mathbf{S}$ confine errors to singular value perturbations rather than singular vector rotations, making them invisible to the polar projection regardless of whether the factors are stored at 4-bit or 8-bit.
We argue that the structural decomposition itself, rather than higher bit-width, is the dominant source of improvement.

\paragraph{Empirical verification.}
We verify this on GPT-2 Small (124M) trained with 1B FineWeb tokens.
For each truncation rank $k$, we compare pure 4-bit quantization against a mixed-precision 8/4-bit scheme (8-bit top factors, 4-bit residual) under two settings: \emph{D only} (decomposition alone) and \emph{Full} (all MuonQ techniques).
Because GRASP uses grid quantization whereas MuonQ uses singular-structure-aligned granularity, we apply MuonQ's column-wise/row-wise aligned quantization to both the 4-bit and 8/4-bit variants, isolating the effect of bit-width from that of granularity.
Table~\ref{tab:grasp_comparison} shows that structural decomposition accounts for nearly all of the PPL gain over Muon4 ($\Delta_\text{total}$), while upgrading the top factors to 8-bit adds little ($\Delta_\text{mixed}$).
Under the Full setting at rank $1/16$, the total improvement is 29.07 and the mixed-precision contribution is only 0.61, about 2\% of the total.
Pure 4-bit Full at this rank (PPL 40.93, 44.3\,MB) also outperforms D-only 8/4-bit (PPL 41.60, 48.1\,MB) in both perplexity and memory, indicating that improving the 4-bit quantizer through companding is more effective than allocating extra bits.
At rank $1/4$, the 8/4-bit variant costs 70.9\,MB, or 87.6\% of Muon8 (81.0\,MB), which largely negates the benefit of low-bit quantization.
Overall, structural decomposition with granularity alignment is the key mechanism, and mixed-precision bit assignment is neither necessary nor cost-effective.

\begin{table}[tbp]
\centering
\vspace{0.5em}
\small
\setlength{\tabcolsep}{6pt}
\begin{tabular}{llcc ccc ccc}
\toprule
 & & & & \multicolumn{3}{c}{\textit{D only}} & \multicolumn{3}{c}{\textit{Full}} \\
\cmidrule(lr){5-7} \cmidrule(lr){8-10}
Method & Bits & Rank & Mem.\ (MB) & PPL $\downarrow$ & $\Delta_\text{total}$ & $\Delta_\text{mixed}$ & PPL $\downarrow$ & $\Delta_\text{total}$ & $\Delta_\text{mixed}$ \\
\midrule
Muon & 32 & --- & 324.0 & 36.36 & --- & --- & 36.36 & --- & --- \\
Muon & 8 & --- & 81.0 & 37.52 & --- & --- & 37.52 & --- & --- \\
Muon & 4 & --- & 40.5 & 70.00 & --- & --- & 70.00 & --- & --- \\
\midrule
MuonQ & 4 & $1/64$ & 41.5 & 49.12 & 20.88 & --- & 42.10 & 27.90 & --- \\
MuonQ & 8/4 & $1/64$ & 42.4 & 47.46 & 22.54 & 1.66 & 42.02 & 27.98 & 0.08 \\
\addlinespace
MuonQ & 4 & $1/16$ & 44.3 & 44.74 & 25.26 & --- & \color{red}{40.93} & 29.07 & --- \\
MuonQ & 8/4 & $1/16$ & 48.1 & \color{red}{41.60} & 28.40 & 3.14 & 40.32 & 29.68 & 0.61 \\
\addlinespace
MuonQ & 4 & $1/4$ & 55.7 & 42.14 & 27.86 & --- & 39.88 & 30.12 & --- \\
MuonQ & 8/4 & $1/4$ & 70.9 & 38.17 & 31.83 & 3.97 & 38.24 & 31.76 & 1.64 \\
\bottomrule
\end{tabular}
\caption{Pure 4-bit vs.\ mixed-precision (8/4-bit) quantization on GPT-2 Small.
\emph{D only}: structural decomposition only. \emph{Full}: normalization + companding + decomposition.
$\Delta_\text{total}$: total PPL improvement over Muon4.
$\Delta_\text{mixed}$: additional improvement from upgrading top-$k$ factors to 8-bit. All variants use MuonQ's singular-structure-aligned granularity.}
\label{tab:grasp_comparison}
\end{table}

\begin{figure}[tbp]
\centering
\begin{subfigure}[b]{0.24\textwidth}
    \includegraphics[width=\textwidth]{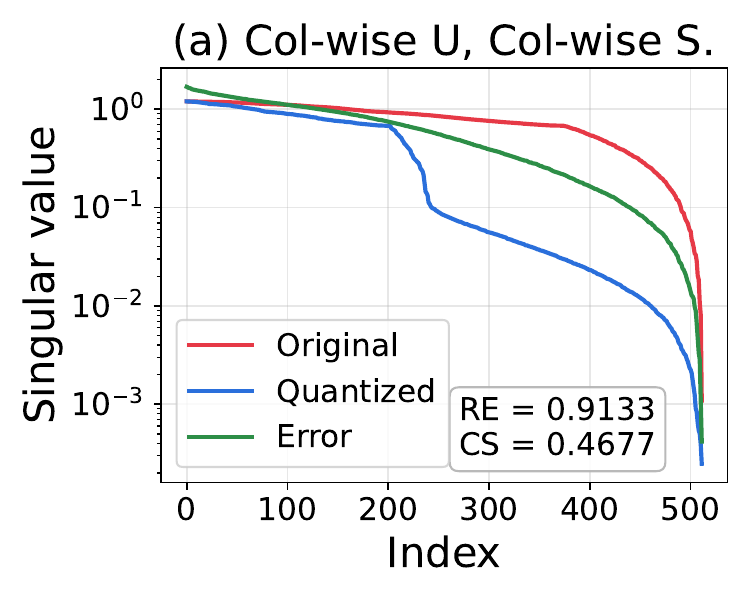}
\end{subfigure}
\hfill
\begin{subfigure}[b]{0.24\textwidth}
    \includegraphics[width=\textwidth]{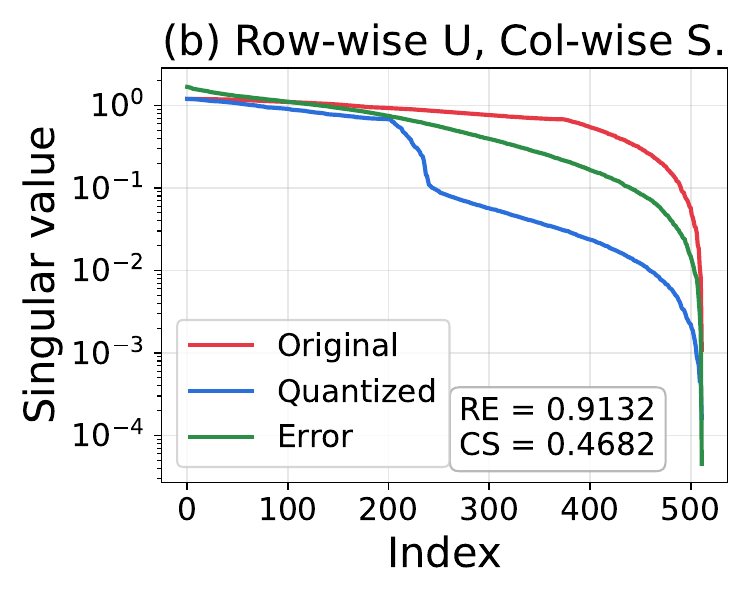}
\end{subfigure}
\hfill
\begin{subfigure}[b]{0.24\textwidth}
    \includegraphics[width=\textwidth]{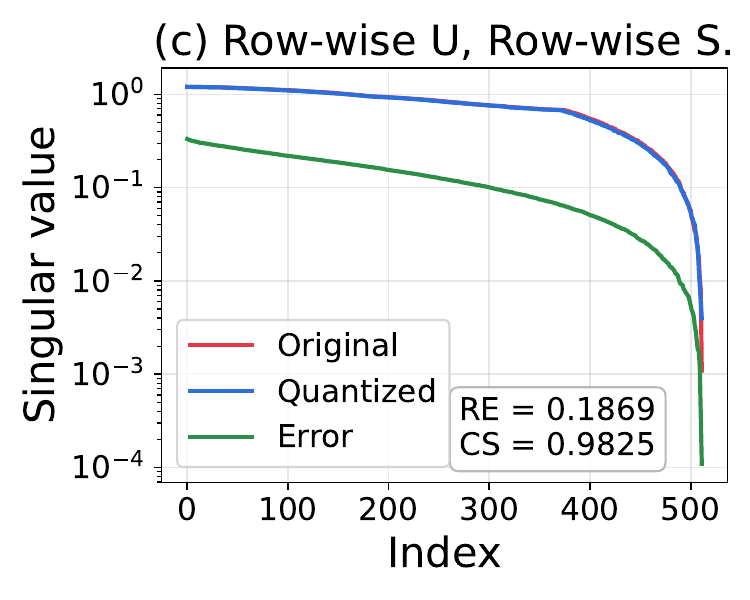}
\end{subfigure}
\hfill
\begin{subfigure}[b]{0.24\textwidth}
    \includegraphics[width=\textwidth]{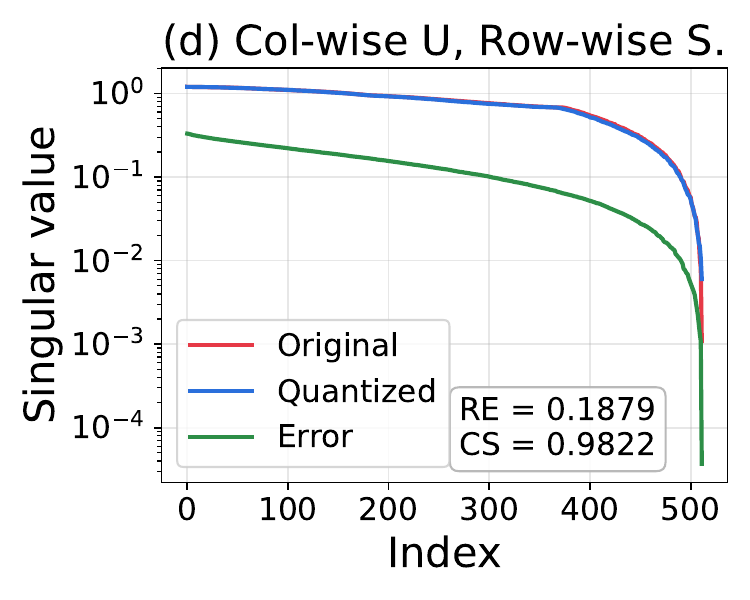}
\end{subfigure}
\caption{Post-orthogonalization singular value spectra under four granularity combinations on \texttt{layers.6.self\_attn.k\_proj}. (a) Col-wise U, Col-wise S. (b) Row-wise U, Col-wise S. (c) Row-wise U, Row-wise S. (d) Col-wise U, Row-wise S.}
\label{fig:granularity_spectra}
\end{figure}

\section{Singular-Structure-Aligned Granularity}
\label{app:granularity_alignment}

As discussed in \S\ref{sec:method-subspace}, MuonQ's structural decomposition requires aligning quantization granularity with the singular structure: column-wise for $\mathbf{U}$ and row-wise for $\mathbf{S}$.
Here we empirically verify that this alignment is critical and that the quantization direction of $\mathbf{S}$ is the dominant factor.

\paragraph{Spectral analysis.}
Figure~\ref{fig:granularity_spectra} shows the post-orthogonalization singular value spectra under four granularity combinations.
The key factor is the quantization direction of $\mathbf{S}$: row-wise quantization of $\mathbf{S}$ (panels c, d) achieves RE $\approx 0.19$ and CS $\approx 0.98$, while column-wise quantization of $\mathbf{S}$ (panels a, b) degrades to RE $\approx 0.91$ and CS $\approx 0.47$.
In contrast, the granularity of $\mathbf{U}$ has negligible effect (comparing a vs.\ b, or c vs.\ d), because $\mathbf{U}$ satisfies $\mathbf{U}^\top\mathbf{U} = \mathbf{I}$ and its columns are already unit-norm and orthogonal, making them inherently robust to the choice of quantization grouping.
This confirms that \textbf{row-wise quantization of $\mathbf{S}$} is the critical design choice.

\paragraph{End-to-end validation.}
\begin{wrapfigure}{r}{0.45\textwidth}
\centering
\vspace{-10pt}
\includegraphics[width=0.45\textwidth]{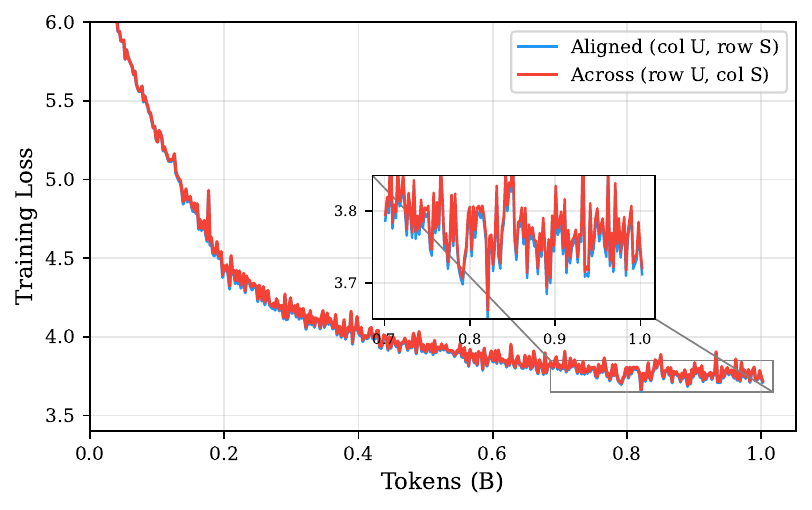}
\caption{Training loss of MuonQ with aligned (col U, row S) vs.\ across (row U, col S) granularity on GPT-2 Small.}
\label{fig:granularity_curve}
\end{wrapfigure}

We further validate this with end-to-end training on GPT-2 Small (124M, 1B FineWeb tokens).
As shown in Figure~\ref{fig:granularity_curve}, the aligned configuration (col-wise $\mathbf{U}$, row-wise $\mathbf{S}$) achieves PPL 40.93, outperforming the across configuration (row-wise $\mathbf{U}$, col-wise $\mathbf{S}$) at PPL 41.30.
Notably, the aligned variant is also \textbf{more memory-efficient}: each quantization group in the aligned scheme requires only $k$ scale factors (one per singular direction), whereas the across scheme requires $m$ or $n$ scale factors (one per matrix row/column), which is significantly larger since $k \ll \min(m,n)$.
Thus, singular-structure alignment improves both accuracy and memory efficiency simultaneously.

\section{$\mu$-law Selection}
\label{app:mulaw}
\begin{wrapfigure}{r}{0.45\textwidth}
\centering
\vspace{-10pt}
\includegraphics[width=0.45\textwidth]{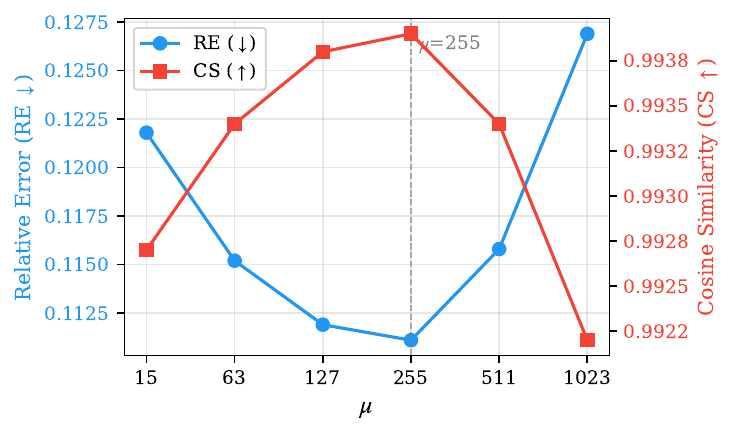}
\caption{Effect of companding parameter $\mu$ on 4-bit row-wise quantization.}
\label{fig:mulaw_selection}
\end{wrapfigure}

The $\mu$-law companding function (Eq.~\ref{eq:mulaw}) contains a single hyperparameter $\mu$ that controls the degree of nonlinear compression.
Following the convention in signal processing, we search over values of the form $2^n - 1$ (i.e., 15, 63, 127, 255, 511, 1023), which correspond to the maximum representable integer under $n$-bit encoding and are the standard choices in ITU-T companding specifications~\citep{mulaw}.
Figure~\ref{fig:mulaw_selection} shows the effect of $\mu$ on 4-bit row-wise quantization quality, measured on \texttt{layers.0.self\_attn.k\_proj}.
Both RE and CS improve as $\mu$ increases from 15 to 255, as the companding function progressively allocates more bins to the dense near-zero region.
Beyond $\mu = 255$, excessive compression distorts the tail values, degrading both metrics.
We select $\mu = 255$ ($= 2^8 - 1$) as the default, which corresponds to the standard ITU-T G.711 $\mu$-law specification and achieves optimal quantization fidelity under row-wise granularity.

\section{Stochastic Rounding}
\label{app:stochastic_rounding}
\begin{wrapfigure}{r}{0.45\textwidth}
\centering
\vspace{-10pt}
\includegraphics[width=0.45\textwidth]{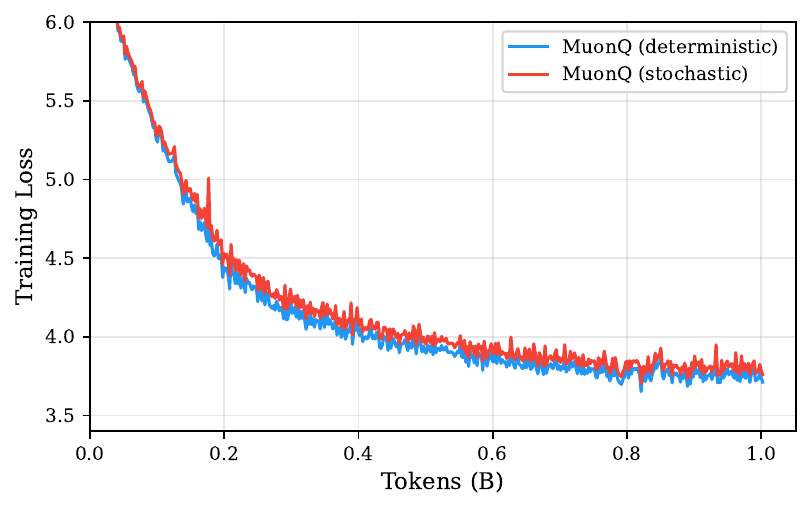}
\caption{Training loss of MuonQ with deterministic vs.\ stochastic rounding on GPT-2 Small. Stochastic rounding degrades performance (42.99 vs.\ 40.93) and increases training variance.}
\vspace{-10pt}
\label{fig:stochastic_rounding}
\end{wrapfigure}
Stochastic rounding is a widely used technique in low-precision training~\citep{gupta2015deep} that replaces the deterministic $\mathrm{round}(\cdot)$ operator with a randomized variant: for a value $z$, it rounds down to $\lfloor z \rfloor$ with probability $\lceil z \rceil - z$ and up to $\lceil z \rceil$ with probability $z - \lfloor z \rfloor$.
Unlike deterministic rounding, stochastic rounding is unbiased in expectation, which has been shown to prevent systematic error accumulation in gradient-based optimization.
Given that MuonQ's pre-quantization normalization (\S\ref{sec:method-normalize}) is also designed to control error accumulation, a natural question is whether stochastic rounding provides complementary benefits.

We evaluate stochastic rounding on GPT-2 Small (124M) with MuonQ at 4-bit, rank $k = \min(m,n)/16$.
As shown in Figure~\ref{fig:stochastic_rounding}, stochastic rounding leads to \emph{worse} training loss than deterministic rounding (final PPL 42.99 vs.\ 40.93), with noticeably higher variance throughout training.
We attribute this to the interaction between stochastic noise and Muon's polar decomposition: since the polar projection discards magnitude information and preserves only directions, the random perturbations introduced by stochastic rounding do not cancel out over time as they would in element-wise optimizers like Adam, but instead inject persistent directional noise into the update.
MuonQ's deterministic rounding combined with pre-quantization normalization provides a more controlled approach that ensures uniform error magnitude without introducing additional stochasticity.

% >>> CR-NEW (camera-ready, Reviewer gTtK W3): full proof of the directional-fidelity bound
{\section{Directional-Fidelity Perturbation Bound}
\label{app:lemma-proof}
We give the derivation of the directional-fidelity bound introduced in \S\ref{sec:methodology}. For $\mat{M}\in\mathbb{R}^{m\times n}$ with quantized reconstruction $\hat{\mat{M}}$ and cosine similarity $\mathrm{CS}(\mat{M},\hat{\mat{M}})=\langle\mat{M},\hat{\mat{M}}\rangle_F/(\|\mat{M}\|_F\|\hat{\mat{M}}\|_F)$,
\begin{equation}
    \|\mathrm{polar}(\hat{\mat{M}})-\mathrm{polar}(\mat{M})\|_F \;\le\; \frac{\sqrt{2}\,\|\hat{\mat{M}}\|_F}{\sigma_{\min}(\mat{M})}\sqrt{1-\mathrm{CS}(\mat{M},\hat{\mat{M}})^2} \;+\; O\!\left(\frac{\|\mat{M}-\hat{\mat{M}}\|_F^2}{\sigma_{\min}(\mat{M})^2}\right).
\end{equation}

\paragraph{Proof sketch.}
Decompose $\hat{\mat{M}}$ into a component parallel to $\mat{M}$ and an orthogonal residual,
\begin{equation}
    \hat{\mat{M}} = \alpha\mat{M}+\mat{E}_\perp, \qquad \alpha=\frac{\langle\mat{M},\hat{\mat{M}}\rangle_F}{\|\mat{M}\|_F^2}, \qquad \langle\mat{M},\mat{E}_\perp\rangle_F=0.
\end{equation}
By the Pythagorean identity in the Frobenius inner product,
\begin{equation}
    \|\mat{E}_\perp\|_F=\|\hat{\mat{M}}\|_F\sqrt{1-\mathrm{CS}(\mat{M},\hat{\mat{M}})^2}.
\end{equation}
Since $\mathrm{polar}(\alpha\mat{M})=\mathrm{polar}(\mat{M})$ for any $\alpha>0$, the parallel component is absorbed by the polar projection and only $\mat{E}_\perp$ contributes to the update error, i.e.\ the effective perturbation is $\Delta\mat{M}=\mat{E}_\perp$. Applying the standard first-order polar perturbation bound with perturbation $\mat{E}_\perp$ then yields the stated inequality.

\paragraph{Consequences.}
The bound has three implications. First, cosine similarity is the \emph{derived} quantization metric for Muon: only the orthogonal component $\mat{E}_\perp$, measured through $\sqrt{1-\mathrm{CS}^2}$, propagates to the update. Second, substituting the bound into existing Muon convergence analyses connects CS to convergence quality. Third, the three components of MuonQ each target a different axis of the same objective of maximizing CS. The constant scales as $1/\sigma_{\min}(\mat{M})$ and can be loose for ill-conditioned $\mat{M}$, so the bound should be read as motivation rather than a worst-case guarantee.}

% >>> CR-NEW (camera-ready, Reviewer uDyi 1.3/1.4): where the decomposition error goes
{\section{Error Contributions of the Top-$k$ and Residual Components}
\label{app:decomp-error}
The structural decomposition (\S\ref{sec:method-subspace}) applies directional treatment (per-vector grouping aligned with the singular structure) to the dominant top-$k$ component while leaving the small-magnitude residual under cheap tensor-wise quantization. Here we quantify why this allocation is well chosen, on the same checkpoint and layer as Figure~\ref{fig:decompose} with $k=\min(m,n)/16$.

\paragraph{The top-$k$ component dominates the pre-polar error budget.}
Table~\ref{tab:decomp-error} reports the pre-polar relative error contributed by each component, measured against $\|\mat{M}_t\|_F$. The top-$k$ component contributes $6.7\times$ the residual's quantization error, so it dominates the error budget. This is by design: the directional treatment is applied precisely to this dominant-error component, so its error is confined to singular-value perturbations and absorbed by the polar projection, while the residual carries only small-magnitude errors.

\begin{table}[htbp]
\centering
\small
\begin{tabular}{lc}
\toprule
Component & Pre-polar RE (relative to $\|\mat{M}_t\|_F$) \\
\midrule
Top-$k$ component ($\mat{U}_k\mat{S}_k$) & 0.2552 \;\textit{(dominant)} \\
Residual ($\mat{R}_k$)                   & 0.0381 \;\textit{(minor)} \\
\bottomrule
\end{tabular}
\caption{Pre-polar quantization-error contributions of the top-$k$ and residual components ($k=\min(m,n)/16$). The top-$k$ component dominates, contributing $6.7\times$ the residual's error.}
\label{tab:decomp-error}
\end{table}

\paragraph{The residual has a substantially smaller numerical range.}
Table~\ref{tab:residual-range} confirms that the residual $\mat{R}_k$ has a much smaller numerical range and smaller singular values than the full momentum $\mat{M}_t$, which is why leaving it under tensor-wise quantization is inexpensive. Together, these measurements show that the benefit of the decomposition comes from aligning the quantizer with the singular structure of the dominant component, not merely from finer per-vector precision.

\begin{table}[htbp]
\centering
\small
\begin{tabular}{lccc}
\toprule
Quantity & $\mat{M}_t$ & $\mat{R}_k$ & Ratio ($\mat{R}_k/\mat{M}_t$) \\
\midrule
$\max|\cdot|$   & 0.0530 & 0.0022 & 4.1\% \\
$\|\cdot\|_F$   & 1.0000 & 0.1148 & 11.5\% \\
$\sigma_1$      & 0.7721 & 0.0239 & 3.1\% \\
\bottomrule
\end{tabular}
\caption{Numerical range and leading singular value of the full momentum $\mat{M}_t$ versus the residual $\mat{R}_k$. The residual is much smaller across all three measures, so tensor-wise quantization of the residual is inexpensive.}
\label{tab:residual-range}
\end{table}}

% >>> CR-NEW (camera-ready, Reviewer uDyi 1.6/2.2): decomposition under a stronger row-wise baseline
{\section{Decomposition under a Stronger Row-wise Baseline}
\label{app:rowwise-baseline}
Granularity (how elements are grouped when computing scale factors) is orthogonal to MuonQ (what is quantized, after what transformation). To show that MuonQ's gains are not an artifact of the tensor-wise baseline used in the main text, we repeat the analysis under a stronger row-wise baseline on GPT-2 Small (124M, 1B FineWeb tokens).

\paragraph{Structural decomposition contributes independently of granularity.}
A direct tensor-wise comparison could suggest that D-only MuonQ4 (PPL 44.7) underperforms a row-wise Muon4 baseline (PPL 42.9). Table~\ref{tab:granularity-matched} shows this is a granularity mismatch: under matched row-wise granularity, D-only MuonQ4 reaches 41.2, improving over row-wise Muon4 (42.9). Decomposition therefore contributes independently of the granularity choice.

\begin{table}[htbp]
\centering
\small
\begin{tabular}{llc}
\toprule
Method & $\mat{R}_k$ granularity & PPL $\downarrow$ \\
\midrule
Tensor-wise Muon4 & ---         & 69.9 \\
Row-wise Muon4    & ---         & 42.9 \\
\midrule
D-only MuonQ4     & tensor-wise & 44.7 \\
D-only MuonQ4     & row-wise    & 41.2 \\
Full MuonQ4       & tensor-wise & 40.9 \\
Full MuonQ4       & row-wise    & \textbf{39.5} \\
\bottomrule
\end{tabular}
\caption{Granularity-matched comparison on GPT-2 Small. Under matched row-wise granularity, D-only MuonQ4 (41.2) improves over row-wise Muon4 (42.9); the earlier apparent contradiction (44.7 vs.\ 42.9) was a granularity mismatch.}
\label{tab:granularity-matched}
\end{table}

\paragraph{Every component retains a positive marginal contribution.}
Table~\ref{tab:rowwise-ablation} repeats the component ablation of Table~\ref{tab:component-ablation} under the row-wise baseline. Each of companding, normalization, and decomposition retains a positive marginal contribution, and full MuonQ4 reaches 39.5 versus 42.9 for row-wise Muon4, confirming that the qualitative pattern of the main ablation is preserved under a different granularity.

\begin{table}[htbp]
\centering
\small
\begin{tabular}{lcccc}
\toprule
Method & \textbf{C} & \textbf{N} & \textbf{D} & PPL $\downarrow$ \\
\midrule
Muon4 baseline (row-wise) & --- & --- & --- & 42.90 \\
\midrule
MuonQ4 & \checkmark & \ding{55}  & \ding{55}  & 42.65 \\
MuonQ4 & \ding{55}  & \checkmark & \ding{55}  & 42.14 \\
MuonQ4 & \ding{55}  & \ding{55}  & \checkmark & 41.20 \\
MuonQ4 & \checkmark & \checkmark & \ding{55}  & 41.49 \\
\midrule
\textbf{Full MuonQ4} & \checkmark & \checkmark & \checkmark & \textbf{39.50} \\
\bottomrule
\end{tabular}
\caption{Component ablation under the row-wise baseline on GPT-2 Small (\textbf{C}: companding, \textbf{N}: normalization, \textbf{D}: decomposition). Each component retains a positive marginal contribution, mirroring Table~\ref{tab:component-ablation}.}
\label{tab:rowwise-ablation}
\end{table}}

% >>> CR-NEW (camera-ready, Reviewer uDyi 3.2/3.3): normalization dynamics + effective-beta
{\section{Effect of Pre-Quantization Normalization on Training Dynamics}
\label{app:norm-dynamics}
Our pre-quantization normalization (\S\ref{sec:method-normalize}) rescales both the gradient and the momentum to unit Frobenius norm at each step. Here we examine, at a larger scale than the GPT-2 Small study in Figure~\ref{fig:norm-ablation}, whether this rescaling alters the optimization dynamics of full-precision Muon, and we make explicit its interpretation as a time-varying momentum coefficient.

\paragraph{Training dynamics are preserved at larger scale.}
We rerun full-precision Muon32 on GPT-2 Medium with and without our normalization. As shown in Figure~\ref{fig:norm-dynamics} (left), the two training-loss curves track each other closely throughout training, with a final-loss gap of only $0.04$, confirming that the normalization introduces no measurable change in optimization dynamics even at this larger scale.

\paragraph{Normalization as a time-varying effective momentum coefficient.}
Our normalization is algebraically equivalent to the standard momentum recursion with a time-varying effective coefficient
\begin{equation}
    \hat{\beta}_t = \beta \cdot \frac{\|\mat{G}_t\|_F}{\|\mat{M}_{t-1}\|_F}.
\end{equation}
Figure~\ref{fig:norm-dynamics} (right) plots $\hat{\beta}_t$ over training. Without normalization, $\hat{\beta}_t$ swings well above the nominal $\beta = 0.95$ early in training before decaying; with normalization it is substantially smoother and smaller. This smoothness is precisely the property that produces uniform per-step quantization error, the condition for isotropic error accumulation established in \S\ref{sec:method-normalize}: the ``dynamic-$\beta$'' view and the ``uniform per-step error'' view describe the same operation, one capturing the algebra and the other why it matters under quantization. The overall smaller $\hat{\beta}_t$ also mildly slows late-training momentum accumulation, which accounts for the small $0.04$ loss gap above.

\begin{figure}[t]
\centering
\includegraphics[width=\textwidth]{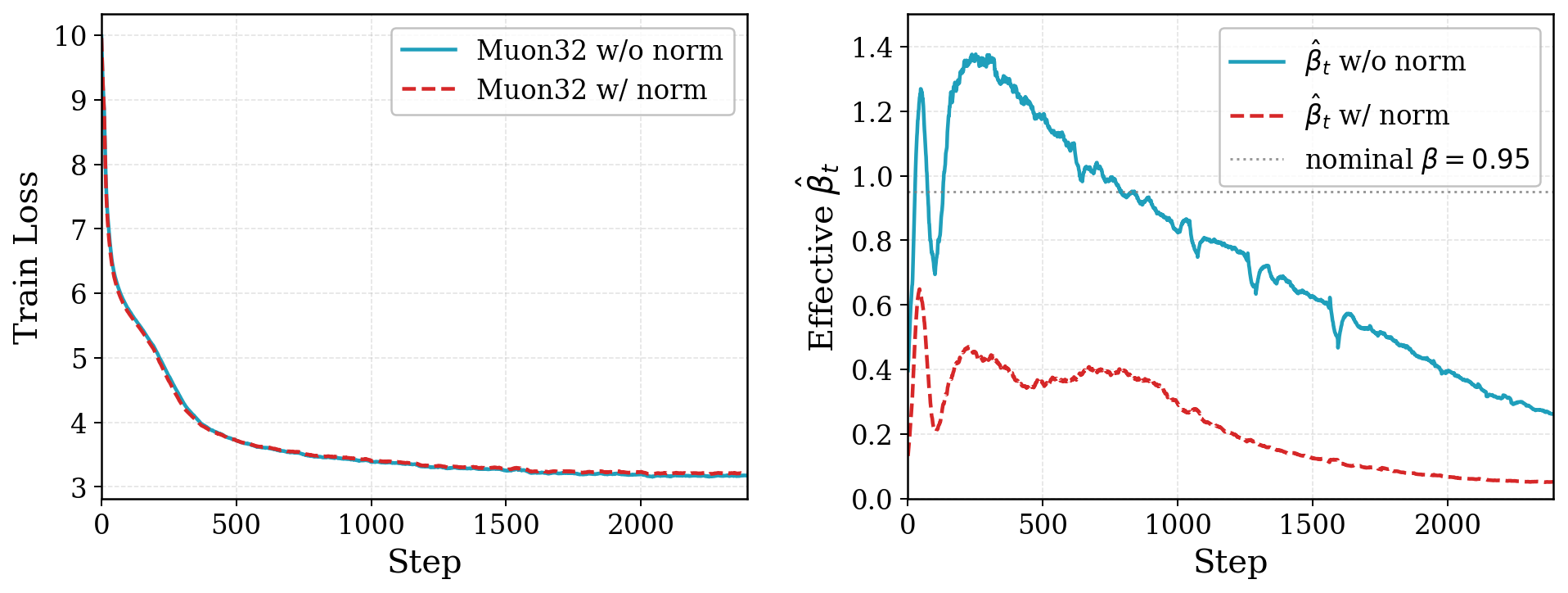}
\caption{Full-precision Muon32 on GPT-2 Medium, with vs.\ without pre-quantization normalization. Left: training loss, with a final-loss gap of $0.04$. Right: effective momentum coefficient $\hat{\beta}_t$; normalization yields a smoother and smaller $\hat{\beta}_t$ relative to the nominal $\beta = 0.95$.}
\label{fig:norm-dynamics}
\end{figure}}

\section{Muon Normalization}
A recurring theme across recent Muon variants is the critical role of normalization applied at different stages of the update pipeline. The standard implementation normalizes the momentum by its Frobenius norm before the Newton--Schulz iteration to ensure convergence of the polar approximation~\citep{jordan2024muon}. Beyond this, NorMuon~\citep{li2025normuon} introduces neuron-wise normalization after orthogonalization to balance per-neuron update magnitudes, while Turbo-Muon~\citep{boissin2025turbomuon} applies a spectral preconditioning step before orthogonalization to accelerate Newton--Schulz convergence. More recently, Muon+~\citep{zhang2026muonplus} demonstrates that a single additional column-row normalization step after the polar factor already yields consistent improvements across model scales. Notably, these methods all improve either efficiency or effectiveness by inserting normalization at different positions, suggesting that Muon is inherently robust to the numerical scale of its input, and that the orthogonal direction, rather than precise magnitude control, is the primary driver of its optimization.
\end{document}